\documentclass[a4paper]{elsarticle}
\usepackage[total={6in, 8in}]{geometry}
\mathchardef\mhyphen="2D

\usepackage{amsmath}
\usepackage{amsthm}
\usepackage{amssymb}
\usepackage{float}
\usepackage[short]{optidef}
\usepackage{subfiles}
\usepackage{subcaption}
\usepackage{bm}
\usepackage{bbm}
\usepackage{tikz}
\usepackage{xfrac}
\usepackage{hyperref}
\usepackage{booktabs}
\usepackage[normalem]{ulem}

\usepackage[shortlabels]{enumitem}
\usepackage[colorinlistoftodos,prependcaption,textsize=tiny]{todonotes}

\newtheorem{theorem}{Theorem}
\newtheorem{corollary}{Corollary}
\newtheorem{lemma}{Lemma}
\newtheorem*{lemma*}{Lemma} 

\DeclareMathOperator{\N}{\mathcal{N}}
\DeclareMathOperator{\Pow}{\text{Pareto}}

\DeclareMathOperator{\E}{\mathrm{E}}
\DeclareMathOperator{\Pb}{\mathrm{P}}

\DeclareMathOperator{\Q}{\mathcal{U}}
\DeclareMathOperator{\V}{\mathcal{V}}

\newcommand\Qgreedy{\Q^{\mathrm{obl}}}
\newcommand\Qdp{\Q^{\mathrm{dp}}}
\newcommand\Qopt{\Q^{\mathrm{opt}}}
\newcommand\Qfair{\Q^{\gamma\mathrm{\mhyphen obl}}}
\newcommand\Qfopt{\Q^{\gamma\mathrm{\mhyphen opt}}}

\newcommand\pxagreedy{\pxa^{\mathrm{obl}}}
\newcommand\pxadp{\pxa^{\mathrm{dp}}}
\newcommand\pxaopt{\pxa^{\mathrm{opt}}}
\newcommand\pxafair{\pxa^{\gamma\mathrm{\mhyphen obl}}}
\newcommand\pxafopt{\pxa^{\gamma\mathrm{ \mhyphen opt}}}

\newcommand\pxbgreedy{\pxb^{\mathrm{obl}}}
\newcommand\pxbdp{\pxb^{\mathrm{dp}}}
\newcommand\pxbopt{\pxb^{\mathrm{opt}}}

\newcommand\pxggreedy{\pxg^{\mathrm{obl}}}

\newcommand\pxgopt{\pxg^{\mathrm{opt}}}

\DeclarePairedDelimiter\ceil{\lceil}{\rceil}
\DeclarePairedDelimiter\floor{\lfloor}{\rfloor}

\newcommand\esp[1]{\mathrm{E}\left[#1\right]}

\newcommand{\w}{W}
\newcommand{\wg}{W_G}

\newcommand{\wh}{\widehat W}
\newcommand{\wt}{\widetilde W}
\newcommand{\wta}{\widetilde{W}_{A}}

\newcommand{\wtg}{\widetilde{W}_{G}}
\newcommand{\wti}{\widetilde{W}_i}
\newcommand{\wha}{\widehat W_A}
\newcommand{\whb}{\widehat W_B}
\newcommand{\whg}{\widehat W_G}

\newcommand{\wi}{W_i}
\newcommand{\whi}{\widehat W_i}

\newcommand{\x}{X_{i|G}}

\newcommand{\pg}{p_G}
\newcommand{\pa}{p_A}
\newcommand{\pb}{p_B}

\newcommand{\phig}{\frac{1}{\sxg}\phi\left(\frac{\hat w-w+\bg}{\sxg}\right)}

\newcommand\alg{{\mathrm{alg}}}
\newcommand\opt{{\mathrm{opt}}}
\newcommand\obl{{\mathrm{obl}}}

\newcommand{\twh}{\hat \theta}
\newcommand{\twha}{\hat\theta_{A}}
\newcommand{\twhb}{\hat\theta_{B}}
\newcommand{\twhg}{\hat\theta_{G}}
\newcommand{\twhgi}{\hat\theta_{G_i}}

\newcommand{\tx}{\twh}

\newcommand{\txg}{\twhg}
\newcommand{\txgdp}{\twhg^{\mathrm{dp}}}

\newcommand{\txggreedy}{\twh^{\mathrm{obl}}}

\newcommand{\txa}{\twha}
\newcommand{\txb}{\twhb}

\newcommand{\sxa}{\sigma_{A}}
\newcommand{\sxb}{\sigma_{B}}
\newcommand{\sxg}{\sigma_{G}}
\newcommand{\sxgi}{\sigma_{G_i}}
\newcommand{\dsx}{\Delta\hat{\sigma}}

\newcommand{\ssgeneric}{\hat{\sigma}}
\newcommand{\ssa}{\hat{\sigma}_{A}}
\newcommand{\ssb}{\hat{\sigma}_{B}}
\newcommand{\ssg}{\hat{\sigma}_{G}}
\newcommand{\ssgi}{\hat{\sigma}_{G_i}}

\newcommand{\stgeneric}{\tilde{\sigma}}
\newcommand{\sta}{\tilde{\sigma}_{A}}
\newcommand{\stb}{\tilde{\sigma}_{B}}
\newcommand{\stg}{\tilde{\sigma}_{G}}
\newcommand{\stgi}{\tilde{\sigma}_{G_i}}

\newcommand{\dst}{\Delta\tilde{\sigma}}

\newcommand{\sqa}{\eta_{A}}
\newcommand{\sqb}{\eta_{B}}
\newcommand{\sqg}{\eta_{G}}
\newcommand{\sqgi}{\eta_{G_i}}

\newcommand{\mqa}{\mu_{A}}
\newcommand{\mqb}{\mu_{B}}
\newcommand{\mqg}{\mu_{G}}
\newcommand{\mqgi}{\mu_{G_i}}
\newcommand{\dmq}{\Delta\mu}

\newcommand{\mq}{\mu}
\newcommand{\sq}{\eta}

\newcommand{\pxa}{x_{A}}
\newcommand{\pxb}{x_{B}}
\newcommand{\pxg}{x_{G}}
\newcommand{\ba}{\beta_A}
\newcommand{\bb}{\beta_B}
\newcommand{\bg}{\beta_G}
\newcommand{\bgi}{\beta_{G_i}}
\newcommand{\db}{\Delta\beta}

\newcommand{\ax}{\alpha}

\newcommand{\amin}{\ax^{\mathrm{min}}}
\newcommand{\amax}{\ax^{\mathrm{max}}}

\newcommand{\ff}{\sfrac{4}{5}}

\begin{document}

\title{On Fair Selection in the Presence of Implicit and Differential Variance 
\footnote{This paper is an extended and revised version of our paper ``On Fair Selection in the Presence of Implicit Variance'' that appeared in the proceedings of EC'20 \cite{emelianov20}.}}

\author[1]{Vitalii Emelianov}
\ead{vitalii.emelianov@inria.fr}

\author[1]{Nicolas Gast}
\ead{nicolas.gast@inria.fr}

\author[2]{Krishna P. Gummadi}
\ead{gummadi@mpi-sws.org}

\author[1]{Patrick Loiseau}
\ead{patrick.loiseau@inria.fr}

\address[1]{Univ. Grenoble Alpes, Inria, CNRS, Grenoble INP, LIG, Grenoble, France}
\address[2]{Max Planck Institute for Software Systems, Saarbrücken, Germany}

\begin{abstract}

Discrimination in selection problems such as hiring or college admission is often explained by implicit bias from the decision maker against disadvantaged demographic groups. In this paper, we consider a model where the decision maker receives a \emph{noisy} estimate of each candidate's quality, whose variance depends on the candidate's group---we argue that such \emph{differential variance} is a key feature of many selection problems. We analyze two notable settings: in the first, the noise variances are unknown to the decision maker who simply picks the candidates with the highest estimated quality independently of their group; in the second, the variances are known and the decision maker picks candidates having the highest expected quality given the noisy estimate. We show that both baseline decision makers yield discrimination, although in opposite directions: the first leads to underrepresentation of the low-variance group while the second leads to underrepresentation of the high-variance group. We study the effect on the selection utility of imposing a fairness mechanism that we term the $\gamma$-rule (it is an extension of the classical four-fifths rule and it also includes demographic parity). In the first setting (with unknown variances), we prove that under mild conditions, imposing the $\gamma$-rule increases the selection utility---here there is no trade-off between fairness and utility. In the second setting (with known variances), imposing the $\gamma$-rule decreases the utility but we prove a bound on the utility loss due to the fairness mechanism.

\end{abstract}


\maketitle

\section{Introduction}
\label{section: introduction}

\paragraph{\textbf{Discrimination in selection and the role of implicit bias}}

Many selection problems such as hiring or college admission are subject to discrimination \cite{Bertrand04a}, where the outcomes for certain individuals are negatively correlated with their membership in salient demographic groups defined by attributes like gender, race, ethnicity, sexual orientation or religion. Over the past two decades, implicit bias---that is an unconscious negative perception of the members of certain demographic groups---has been put forward as a key factor in explaining this discrimination \cite{implicit_bias_greenwald06}. While human decision makers are naturally susceptible to implicit bias when assessing candidates, algorithmic decision makers are also vulnerable to implicit biases when the data used to train them or to make decisions was generated by humans.

To mitigate the effects of discrimination on candidates from underrepresented groups, various fairness mechanisms\footnote{These mechanisms are sometimes termed ``positive discrimination'' (e.g., in Germany, France, China, or India) or ``affirmative actions'' (in the USA), often referring to their justification as corrective measures against discrimination suffered in the past by disadvantaged groups. In our work, we analyze the effect of these mechanisms in a particular setting of selection problems (with differential variance) independently of their motivation, hence we use the more neutral term ``fairness mechanisms.''} are adopted in many domains, either by law or through softer guidelines. For instance, the \emph{Rooney rule} \cite{rooney_rule_collins07} requires that, when hiring for a given position, at least one candidate from the underrepresented group be interviewed. The Rooney rule was initially introduced for hiring American football coaches, but it is increasingly being adopted by many other businesses in particular for hiring top executives \cite{Cavicchia15a,Passariello16a}. Another widely used fairness mechanism is the so-called \emph{$\ff$-rule} \cite{holzer00}, which requires that the selection rate for the underrepresented group be at least 80\% of that for the overrepresented group (otherwise one says that there is adverse impact). This rule is part of the ``Uniform Guidelines On Employee Selection Procedures''\footnote{A set of guidelines jointly adopted by the Equal Employment Opportunity Commission, the Civil Service Commission, the Department of Labor, and the Department of Justice in 1978.}. A stricter version of the $\ff$-rule is the so-called \emph{demographic parity} constraint, which requires the selection rates for all groups to be equal. An overview of these and other fairness mechanisms can be found in \cite{holzer00}.

Fairness mechanisms, however, have been the subject of frequent debates. On the one hand, they are believed to promote the inclusion of deserving candidates from underrepresented groups who would have otherwise been excluded in particular due to implicit bias. On the other hand, they are viewed as requiring consideration of candidates from underrepresented groups at the expense of candidates from overrepresented groups, which may potentially decrease the overall utility of the selection process, i.e., the overall quality of selected candidates.

\paragraph{\textbf{Formal analysis of fairness mechanisms in the presence of implicit bias}}

Perhaps surprisingly, the mathematical analysis of the effect of fairness mechanisms on utility in the context of selection problems was initiated only recently by Kleinberg and Raghavan \cite{kleinberg18} (see also an extension to ranking problems in \cite{celis20}). The authors of \cite{kleinberg18} assume that each candidate $i$ has a true latent quality $\wi$ that comes from a group-independent distribution. They model implicit bias by assuming that the decision maker sees an estimate of the quality $\whi = \wi$ for candidates from the well-represented group and $\whi = \wi / \beta$ for candidates from the underrepresented group, where $\beta > 1$ measures the amount of implicit bias. The factor $\beta$ is unknown (as it is implicit bias) and the decision maker selects candidates by ranking them according to $\whi$. Then Kleinberg and Raghavan \cite{kleinberg18} show that, under a well-defined condition (that roughly qualifies scenarios where the bias is large), the Rooney rule improves in expectation the utility of the selection (measured as the sum of true qualities of candidates selected for interview). This result contradicts conventional wisdom that fairness considerations in a selection process are at odds with the utility of the selection process. Rather, it formalizes the intuition that, in the presence of strong implicit bias (which makes it hard to compare candidates across groups), considering the best candidates across a diverse set of groups not only improves fairness but it also has a positive effect on utility.

\paragraph{\textbf{The phenomenon of differential variance and its role in discrimination}}

In this paper, we identify and analyze a fundamentally different source of discrimination in selection problems than implicit bias. Even in the absence of implicit bias in a decision maker's estimate of candidates' quality, the estimates may differ between the different groups in their \emph{variance}---that is, the decision maker's ability to precisely estimate a candidate's quality may depend on the candidate's group. There are at least two main reasons for group-dependent variances in practice. The first arises from \emph{candidates}: different groups of candidates may exhibit different variability when their quality is estimated through a given test. For instance, students of different genders have been observed to show different variability on certain test scores \cite{gender_variability_baye16,O'Dea18a}.\footnote{Note that, while this indicates that observed performance is more variable for one group than the other, it is impossible to tell whether this comes from different underlying distributions or from different measurement variances---or (more likely) from both. In fact, the general ``variability hypothesis'' is subject to a number of controversies. Nevertheless, this indicates potential differences between groups in the variance of the observed signals and our model can flexibly incorporate both different prior distributions and different measurement noises.} The second arises from the \emph{decision makers}: decision makers might have different levels of experience (or different amounts of data in case of algorithmic decision making) judging candidates from different groups and consequently, their ability to precisely assess the quality of candidates belonging to different groups might be different. For instance, when hiring top executives, one may have less experience in evaluating the performance of female candidates because there have been fewer women in those positions in the past (in France for instance, there was only one woman CEO amongst the top-40 companies in 2016-2020 \cite{Lexpress20a}). The quality estimate's variance might also change from one decision maker to another. For example, in college admissions, recruiters might be able to judge candidates from schools in their own country more accurately than those from international schools.

We refer to the above phenomenon as \emph{differential variance} as the variance of the quality estimate is group-dependent. We posit that differential variance is an omnipresent and fundamental feature affecting selection problems (including in algorithmic decision making). Indeed, having different variances for the different groups is mostly inevitable and hardly fixable. 
In this paper, we model the differential variance phenomenon by assuming that the decision maker sees of an estimate of the quality of a candidate $\whi$ that is equal to the candidate's true latent quality $\wi$ (possibly with an additional bias term) plus an additive noise\footnote{This noise may be a property of the decision maker getting a noisy perception of the candidate's quality or a property of the candidate (i.e., the variability in the candidate's performance).}  whose variance depends on the group of the candidate. 

%


We distinguish between two notable settings. In the first setting, the noise variance is assumed to be unknown to the decision maker---we then call it \emph{implicit variance}. In this case, a natural baseline decision maker is the \emph{group-oblivious} algorithm\footnote{Throughout the paper, we use the term `algorithm' for the selection procedure, irrespective of whether it is algorithmic decision making or not.} that simply selects the candidates with the highest estimated quality $\whi$, irrespective of their group. The group-oblivious selection algorithm can represent not only a decision maker unaware of the implicit variance in their estimates, but also a decision maker determined to not use group information---as it may be the case for instance in college admission based on standardized tests. In the second setting, the noise variance is known to the decision maker. In this case, a natural baseline is the \emph{Bayesian-optimal} algorithm: this decision maker can use the group information as well as the knowledge of the distributions of latent quality and noise to select the candidates that maximize the expected quality given the noisy estimate.

As a first cornerstone, our analysis shows that, in the presence of differential variance, both the group-oblivious and the Bayesian-optimal algorithms lead to discrimination (although in opposite directions, see the overview of our results below). A natural way to address this representation inequality is to adopt fairness mechanisms proposed to address discrimination in selection such as the ones discussed above; but this poses the same question that was investigated by Kleinberg and Raghavan \cite{kleinberg18} in the case of implicit bias: \emph{what is the effect of fairness mechanisms on the quality of a selection in the presence of differential variance?}


\paragraph{\textbf{Our model and overview of our results}}

To answer this question, we propose a simple model with two groups of candidates $A$ and $B$: for each candidate $i$, the decision maker receives a noisy (and possibly biased) quality estimate $\whi = \wi -\bgi + \sxgi \varepsilon_i$, where $G_i$ is the group to which the candidate belongs and $\varepsilon_i$ is a standard normal random variable. The estimator has an additive bias $\bgi$ and a variance $\sxgi^2$ that depend on the candidate's group. We assume that the true quality $\wi$ comes from a distribution---assumed normal in our analytical results---that may be group-dependent. 
The decision maker then selects a fraction $\ax$ (called selection budget) of the candidates. 

The key feature of our model is the variance $\sxgi^2$ that depends on the candidate's group---to model differential variance. In its general version, we also allow a bias and a latent quality distribution that depend on the candidate's group. Using this general model, we first show (Section~\ref{ssec:one-stage:policies}) that both the group-oblivious and the Bayesian-optimal selection algorithms systematically lead to underrepresentation---i.e., lower selection rate---of one of the groups of candidates. Specifically, we identify a cutoff budget such that the group-oblivious selection algorithm leads to underrepresentation of the low-variance group for any budget $\ax$ smaller than the cutoff (the most common case) and underrepresentation of the high-variance group for any budget $\ax$ larger than the cutoff. Conversely (and for a different cutoff), the Bayesian-optimal algorithm leads to underrepresentation of the high-variance group for low budgets and of the low-variance group for high budgets. In fact, we show (Section~\ref{ssec:group-independent quality}) that this is true even in the absence of bias and with group-independent latent quality distributions---that is, if the noise variance is the only thing that depends on the candidate's group. In this particular case, the cutoff budget for both algorithms is $\ax=1/2$.

Then we investigate how the utility of the group-oblivious and the Bayesian-optimal baselines are affected when imposing a fairness mechanism. Specifically, we study a generalization of the $\ff$-rule that we call \emph{$\gamma$-rule}, which imposes that the selection rate for a given group is at least $\gamma$ times that of the other group for some parameter $\gamma \in [0, 1]$. This includes both the $\ff$-rule ($\gamma=0.8$) and demographic parity ($\gamma=1$) as special cases. In the general model, we identify conditions under which the $\gamma$-rule never decreases the utility of the group-oblivious algorithm (Section~\ref{ssec:improve})---that is, there is no trade-off between fairness and selection quality for this baseline. The utility even strictly increases for $\gamma$ close enough to one, including for demographic parity. Interestingly, in the special case without bias and with group-independent latent quality distributions---that is, with only implicit differential variance---, this result \emph{always} holds for any parameters (Section~\ref{ssec:group-independent quality}). Compared to the Bayesian-optimal baseline, the $\gamma$-rule cannot increase the utility (since Bayesian-optimal is already optimal given the available information). We prove, however, a bound on the ratio of the utility of the Bayesian-optimal algorithm with and without the $\gamma$-rule imposed, which limits the decrease of utility due to imposing a fairness mechanism in this setting. Our bound is valid in the general model (Section~\ref{ssec:harm}) but takes a particularly simple form in the special case without bias and with group-independent latent quality distributions (Section~\ref{ssec:group-independent quality}).

A typical case of differential variance is when the decision maker has more uncertainty about one group, due to lack of statistical confidence (e.g., in hiring). In such a case, the high-variance group naturally corresponds to the minority group. The group-oblivious algorithm would then overrepresent the minority group (for small selection budgets), and the fairness mechanism would lead to selecting fewer of the minority group---which is counter-intuitive. We stress, however, that those are typically cases where the relevant baseline is the Bayesian-optimal algorithm, which behaves very differently. Through the Bayesian posterior quality computation, this baseline would disregard candidates for which the observed quality estimate is uninformative, that is the high-variance group. As mentioned above, we indeed find that the Bayesian-optimal algorithm underrepresents the high-variance group (i.e., the minority), and that the fairness mechanism increases the proportion of selected high-variance candidates---which is coherent with intuition for that case. The group-oblivious baseline is meaningful in other scenarios, typically when the decision maker is not allowed to use the group information (e.g., in college admission based on standardized tests). In such cases, the high-variance group may not be a minority group (and our model does not require that it is). 

At a high-level, our results indicate that, with differential variance, the two decision makers (group-oblivious and Bayesian-optimal) lead to nearly opposite outcomes in terms of discrimination; and that the effect of imposing fairness mechanisms can be very different for both. These results imply that a policy-maker considering fairness mechanisms for a given problem should first evaluate to which decision maker the selection rule corresponds, and then choose whether or not to recommend the $\gamma$-rule based on it. Note that this should be fairly easy to distinguish between the two in practice, since one conditions on group identity while the other does not.

\paragraph{\textbf{Organization of the paper}}

The rest of the paper is organized as follows. We present the model in Section~\ref{section: model}. We give all the results in the most general case in Section~\ref{section: one-stage general case}. Due to their generality, those results are sometimes complex. In Section~\ref{section: cases}, we analyze three notable cases for which the results are easier to interpret: the case without bias and with group-independent latent quality distributions (Section~\ref{ssec:group-independent quality}), the case with bias but with group-independent latent quality distributions (Section~\ref{ssec:bias}), and the case without bias but with group-dependent latent quality distributions (Section~\ref{ssec:group-dependent}). Through numerical simulations in  Section~\ref{section: experiments}, we extend our analytical results, in particular to cases where the latent quality distribution does not follow a normal law. We conclude in Section~\ref{section: discussion}.

\paragraph{\textbf{Related work}}

There is an abundant literature on fairness in machine learning, in particular on classification, that tackles the question of how to learn a classifier while enforcing some fairness notion in the outcome \cite{Pedreshi08a,Hardt:2016,Zafar17c,Zafar17a,Chouldechova17a,Corbett-Davies:2017,Lipton18a,Mathioudakis19a}. In this literature, fairness is usually seen as a constraint that reduces the classifier's accuracy and the fairness-accuracy tradeoff is analyzed. In contrast, in our work, we examine selection problems in which fairness can improve utility. Selection also differs from classification by the presence of selection budgets (i.e., maximal number of class-1 predictions), which changes the problem significantly.

The problem of selection is considered in \cite{kleinberg18} under the presence of implicit bias \cite{implicit_bias_greenwald06}. In their work, the authors study the Rooney rule \cite{rooney_rule_collins07} as a fairness mechanism and show that under certain conditions, it improves the quality of selection. An extension of the Rooney rule is studied under a similar model in \cite{celis20}, where the authors investigate the ranking problem (of which the selection problem can be seen as a special case) also in the presence of implicit bias and obtain similar results. In both papers, simple mathematical results expressing conditions under which the Rooney rule improves utility are obtained in the limit regime where the number of candidates is very large; we use the same limit regime in our work. In contrast to those papers that only consider bias, we introduce in addition the notion of differential variance to capture the difference in precision of the quality estimate for different groups. We also consider an additive bias rather than a multiplicative one as it makes more sense for normally distributed qualities. Although our model incorporates both an additive bias and differential variance (in Section~\ref{section: one-stage general case}), we purposely restrict it in Section~\ref{section: cases} to the simplest possible form of differential variance so as to show its effect on the selection problem independently of bias. In our work, we also consider the $\ff$-rule \cite{holzer00} (or rather an extension of it that we call the $\gamma$-rule and that includes demographic parity) rather than the Rooney rule. The main difference between the two is that the $\ff$-rule imposes a constraint on the \emph{fraction} of selected candidates from the underrepresented group whereas the Rooney rule or its extension in \cite{celis20} imposes a constraint on the \emph{number} of selected candidates from the underrepresented group.

Implicit bias, or simply bias (possibly from an algorithm trained on biased data) in the evaluation of candidates quality is certainly a primary factor of discrimination; but it is also one that may reasonably be fixable through the use of algorithms combined with appropriate debiasing techniques and ground truth data \cite{Raghavan20a} (e.g., by learning fair representations of data \cite{zemel13, locatello19}). The effects of bias can be also fixed by introducing some fairness constraints on learned prediction models. For example, in \cite{wick2019}, the binary classification problem in the presence of label bias is studied and it is shown that adding a demographic parity constraint to an empirical risk minimization  problem can lead to better generalization.  Similarly, in \cite{blum2019recovering}, the authors study the effects of label bias on binary classification and they show that equal opportunity  fairness criterion (that ensures that true positives are equal across the groups) can reduce the bias in prediction for most of the reasonable cases, as well as improve the accuracy of classification. In \cite{dutta_icml2020}, the authors quantify a fairness-accuracy trade-off using an information theoretic approach and, in addition, they show that for the majority of traditional fairness criteria (like equal opportunity and demographic parity) there exists an ideal data distribution for which fairness and Bayesian optimality are in accordance.

The notion of differential variance first appeared (with different terminology) in the seminal work of Phelps \cite{phelps72} to explain racial inequality in wages. There, a Bayesian decision maker observes noisy signals of productivity of each worker. Productivities are assumed to be drawn from a common distribution while precisions of estimation differ across races. Phelps shows that a Bayesian decision maker that assigns wages equal to the expected productivity of a worker leads to inequality of wages: in the region of high values of signals the low-precision workers receives lower wages. Our model is similar that of Phelps, with additional bias and possibly group-dependent prior distributions. We also study cases where the variance is implicit---hence the decision maker cannot use Bayes' rule to estimate expected quality given noisy estimates---, and focus on utility for our main results.

This paper is an extended version of our paper ``On Fair Selection in the Presence of Implicit Variance'' \cite{emelianov20}. We extend it by considering the general model with bias and group-dependent latent quality distributions, and by analyzing in parallel the two baselines of the group-oblivious and Bayesian-optimal algorithms (whereas \cite{emelianov20} only looks at the group-oblivious baseline, that is at implicit variance). On the other hand, we do not include the results on two-stage decisions makers for conciseness. Following \cite{emelianov20}, Garg et al. \cite{garg21} studied a similar model (using the term differential variance that we also adopt here). The authors propose a model of school admission with students of two groups: advantaged and disadvantaged. Each student has an intrinsic quality which is not observable to schools: only noisy signals of the quality are available. The advantaged and disadvantaged students differ in the level of precision of their signals, and can also differ in their ability to access the tests. The authors consider the case of a Bayesian school of limited capacity. They study how different policies made by the school (group-aware and group-unaware) affect the diversity level, individual fairness and overall merit of admitted students. The authors also study how dropping test scores and different abilities to access tests affects the above characteristics.

Fairness mechanisms has been a subject of a number of studies in the economic literature, in particular from empirical data. In \cite{coate93}, the authors study whether affirmative actions can remove stereotypes about a particular population. In \cite{aff_action_balafoutas12}, an empirical  evaluation of the influence of affirmative actions in recruiting is performed and it is shown that it can bring quality together with equality. Our work complements those studies through a theoretical model that leads to analytical results on the effect of fairness mechanisms in the presence of differential variance.


\section{Model and selection algorithms}
\label{section: model}

\subsection{The model of selection with differential variance}

We consider the following scenario. A decision maker is given $n$ candidates, out of which a subset of size $m=\alpha n$ is selected, $\alpha \in(0,1)$. We assume that the set of candidates can be partitioned in two groups: group $A$ and group $B$. There are $n_A$ candidates from group $A$ and $n_B=n-n_A$ candidates from group $B$.  We refer to them as $A$-candidates and $B$-candidates.

Each candidate $i\in\{1, \dots, n\}$ is endowed with a true latent quality $\wi$.  We assume that the qualities $\wi$ are drawn \emph{i.i.d.} from an underlying probability distribution that can be group-dependent.\footnote{We present here the model in its most general form. We will analyze special cases, in particular when the  quality distribution is group-independent, in Section~\ref{section: cases}.}  For our analytical results, we assume that this distribution is a normal distribution of mean $\mqgi$ and variance $\sqgi^2$, where $G_i\in\{A, B\}$ is the group of candidate~$i$.

The goal of the decision maker is to maximize the expected quality of the selected candidates: $\esp{\sum_{i\in\text{selection}} \wi}$.
When making the selection decision, the decision maker has access to a (possibly biased) noisy estimator of the true quality. We denote the estimator of the quality of candidate $i$ by $\whi$. We assume that the bias and the variance of the estimator may depend on the group: for a candidate $i$ that belongs to group $G_i\in\{A,B\}$, its estimated quality is
\begin{equation}
  \label{eq: x1g}
  \whi = \left\{
    \begin{array}{ll}
      \wi - \ba + \sigma_{A} \cdot \varepsilon_i & \text{ if $i$ is an $A$-candidate,}\\
      \wi - \bb + \sigma_{B} \cdot \varepsilon_i &\text{ if $i$ is a $B$-candidate,}
    \end{array}
    \right.
\end{equation}
where $\varepsilon_i$ is a centered random variable from $\N(0,1)$---the standard normal distribution, of mean $0$ and variance $1$. The variables $\varepsilon_i$ are assumed independent and identically distributed.  Note that we model the bias as an additive parameter in contrast to the multiplicative parameter in \cite{kleinberg18}. This is more suitable for our model of qualities as normally distributed random variables, which can be negative (a multiplicative bias on a negative quality would turn into a positive effect, which is not meaningful). 

We denote by $\ssgi^2=\sxgi^2 + \sqgi^2$ the variance of the estimate $\whi$. Without loss of generality, we assume that the estimates' variance is larger for $A$-candidates than for $B$-candidates, that is $\ssa^2 > \ssb^2$. We note that none of our results require that $A$ is also the minority group, i.e., that $n_A < n_B$. It is possible to think of scenarios where the minority group has lower variance in cases where the difference in variances arises from the candidates. In the example of students' tests scores (see Section~\ref{section: introduction}), for instance, one could potentially observe that males have greater variability in topics in which they are in majority. If the difference in variances arises from the decision maker and has a statistical nature, the minority group (for past selections) will have higher variance due to less data points to build the estimator.

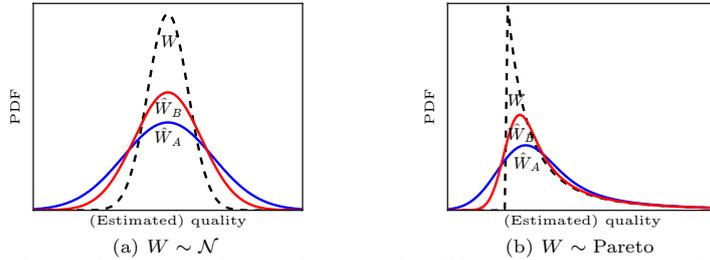
\begin{figure}
  \centering
  \begin{subfigure}{0.35\textwidth}
  \centering
  \begin{tikzpicture}
    \node (img)  {\includegraphics[scale=0.45]{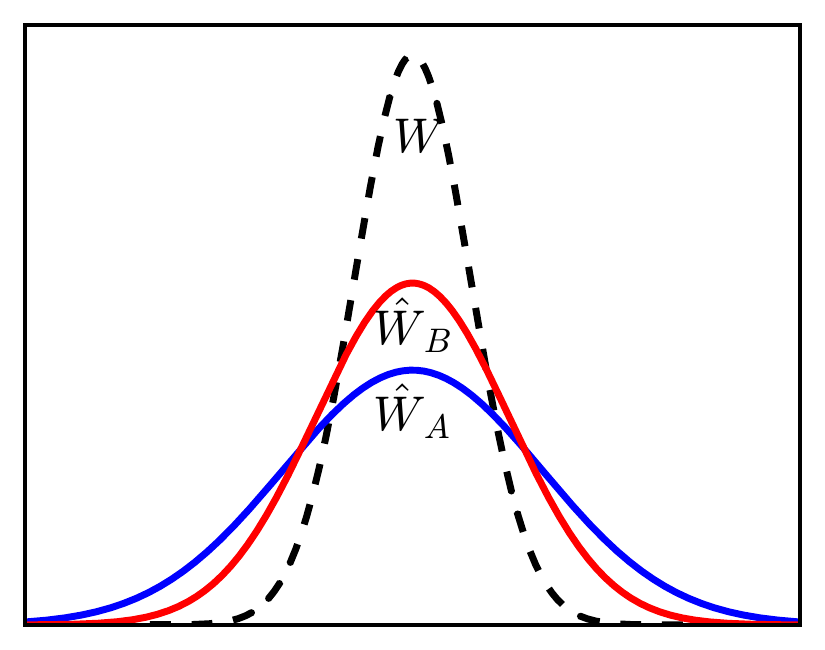}};
    \node[below=of img, node distance=0cm, yshift=1.3cm] {\tiny (Estimated) quality};
    \node[left=of img, node distance=0cm, rotate=90, anchor=center,yshift=-1cm] {\tiny PDF};
   \end{tikzpicture}
   \vspace{-.2cm}\caption{$\w \sim \N$}
  \end{subfigure}
  \begin{subfigure}{0.35\textwidth}
    \centering
      \begin{tikzpicture}
        \node (img)  {\includegraphics[scale=0.45]{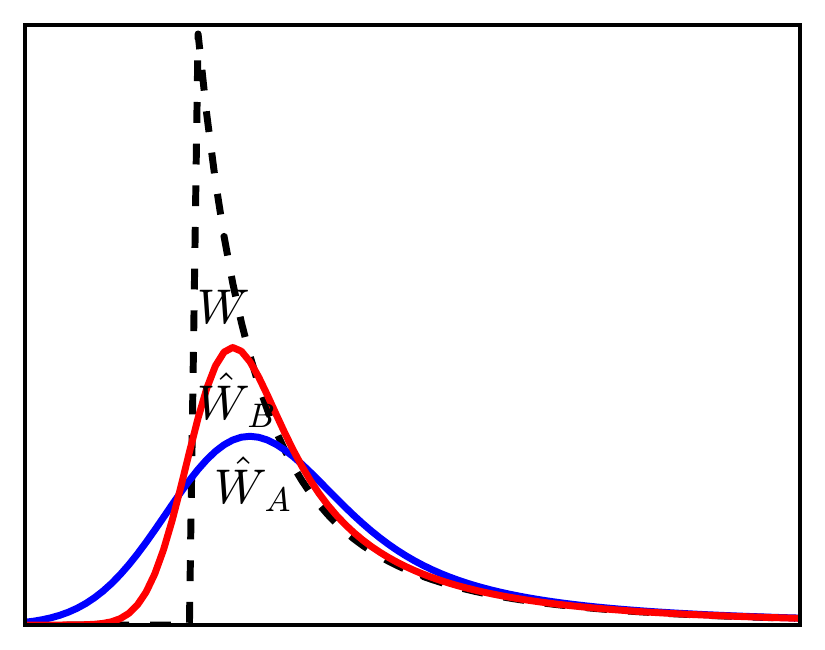}};
        \node[below=of img, node distance=0cm, yshift=1.3cm] {\tiny (Estimated) quality};
        \node[left=of img, node distance=0cm, rotate=90, anchor=center,yshift=-1cm] {\tiny PDF};
      \end{tikzpicture}
      \vspace{-.2cm}\caption{$\w \sim \Pow$}
    \end{subfigure}
\vspace{-.4cm}\caption{Probability density function of the true latent quality $\w$ and the estimated quality $\wh$. To the purpose of illustration, the underlying distribution is assumed group-independent and the estimation is unbiased.}
\vspace{-.4cm}
\label{fig: pdf illustration}
\end{figure}

Throughout the paper, we refer to this difference in variance as \emph{differential variance} because we assume that the variance of the estimators differs across groups. Fig.~\ref{fig: pdf illustration} illustrates the resulting distribution of quality estimates for groups $A$ and $B$ for different distributions of the true latent quality (by abuse of notation, we denote by $\wha$ a variable that has the same distribution as $\wi + \sigma_{A} \varepsilon_i$ and similarly for $B$).


\subsection{Selection algorithms}
\label{sec:selection_algorithms}

Candidates are selected in a one-stage
process: for each candidate $i$, the decision maker
observes the quality estimate $\whi$ as well as its group
$G_i\in\{A,B\}$. The decision maker then selects $m$ candidates out of those $n$.  The goal of the decision maker is to maximize the expected quality of the selected $m$ candidates. 
In this paper, we distinguish and study the following two baseline selection algorithms. Each baseline is a natural selection algorithm in a situation when the decision maker knows the parameters of the model ($\mqgi$, $\sqgi^2$, $\bgi$ and $\sxgi^2$) or not. 

\paragraph{Group-Oblivious Algorithm}
One of the most natural selection rules is to sort the candidates according to $\whi$ irrespective of their group and to keep the best $m$. We call this the \emph{group-oblivious} selection algorithm. Typical examples of the group-oblivious algorithm could be admission processes in colleges where the selection is performed with respect to standardized test results (no group information is taken into account), or selection processes where the decision maker does not know the model's parameters, and in particular where it does not know the variance of the estimator (hence the name \emph{implicit variance} in that case).   This selection algorithm might be also seen  as a fair treatment because the selection does not use the group label. Yet, because of the differential variance or bias, this might lead to discrimination. We will discuss that in Theorem~\ref{theorem: group oblivious general case}.

\paragraph{Bayesian-Optimal Algorithm} When the variance of the noise is known, an alternative selection algorithm is what we call the \emph{Bayesian-optimal} algorithm. This algorithm knows all the parameters of the problem (the quality distribution, the variances of noise $\sxg^2$, and the biases $\bg$) and chooses the candidates with the largest expected quality given the estimate $\whi$. 
Since $(\wi, \whi)$ is a bivariate normal random vector, then using the property of conditional expectation for normal random vectors, the expected quality of candidate given its quality estimate can be expressed as:
\begin{equation}
  \wti = \E(\wi|\whi) = \frac{\sqgi^2}{\sxgi^2 + \sqgi^2}(\whi + \bgi) + \left(1 - \frac{\sqgi^2}{\sxgi^2 + \sqgi^2} \right) \mqgi.
\label{eq:expected w given w hat}
\end{equation}
Note  that  $\wti$ converges to $\whi + \bgi$ as $\sxgi^2$ tends to $0$ (i.e., there is no noise) and it converges to $\mqgi$ as $\sxgi^2$ tends to $\infty$. Intuitively, all candidates appear similar to the decision maker as the precision of estimation degrades. 
We denote by $\stgi^2 = \sqgi^4 / \left(\sqgi^2 + \sxgi^2\right)$ the variance of the expected quality $\wti$.

\begin{figure}
  \centering
  \includegraphics[width=0.9\textwidth]{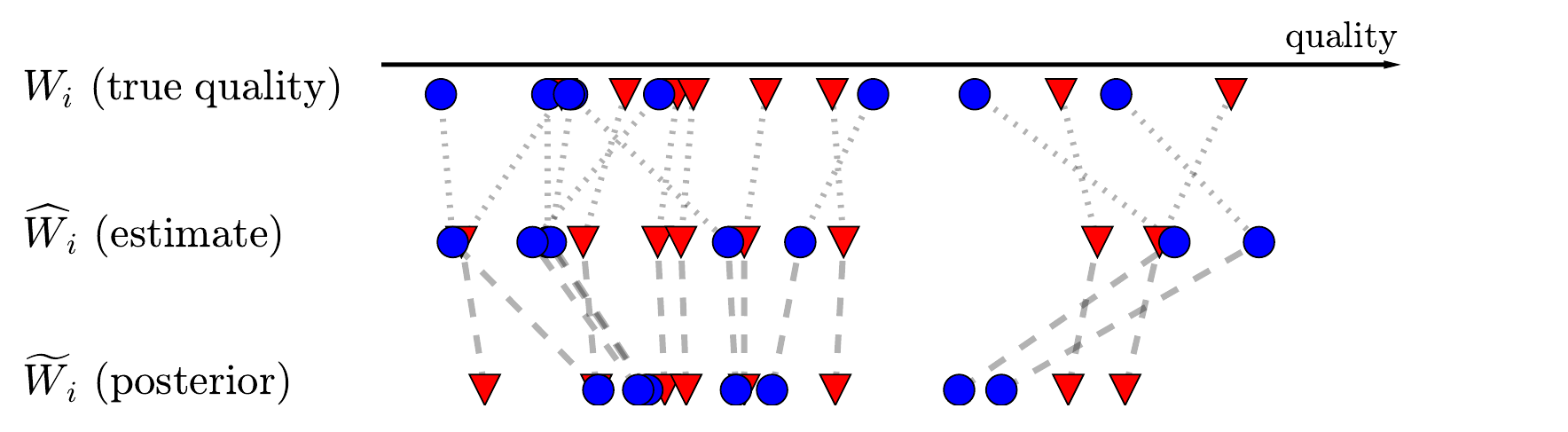}
  \caption{Illustration of the baseline selection algorithms. Here, there are $n_{blue}=8$ blue and $n_{red}=8$ red candidates, and the decision maker wants to select $m=2$ candidates. The quality is group-independent and there is no bias. The estimator variance is three times higher for the blue candidates. Here, the group-oblivious algorithm would select the $2$ blue candidates. Yet, because blue candidates have higher variance, the Bayesian-optimal algorithm would select $2$ red.}
  \label{fig:illustration}
\end{figure}

Perhaps more surprisingly, the Bayesian-optimal algorithm also leads to discrimination (although in the opposite way as for the group-oblivious algorithm) as we show in Theorem~\ref{theorem: bayesian optimal general case}. 
We illustrate how the two decision making algorithms work with an example depicted in Figure~\ref{fig:illustration}. In this example, the blue candidates have higher variance $\sigma_{blue}=3\sigma_{red}$. This implies that the posteriors $\wti$ are more shrank towards the mean for blue candidates than for red candidates: as a result, the Bayesian-optimal tends to select fewer blue candidates compared to red candidates. Note that the Bayesian-optimal is only optimal \emph{in expectation} given the information available; it needs not be optimal for a given realization (on Figure~\ref{fig:illustration} the optimal selection ex-post would be one red and one blue).

  

\subsection{The $\gamma$-rule fairness mechanism}
\label{ssec:fairness mechanisms}


For a given algorithm $\alg\in\{\obl,\opt\}$, we denote by $\pxa^\alg$ (and $\pxb^\alg$) the proportion of the $A$-candidates (and $B$-candidates) that are selected, where $\obl$ stands for group-oblivious and $\opt$ for Bayesian-optimal.  A selection algorithm might favor one group or the other, that is $\pxa^\alg \gg \pxb^\alg$ or $\pxb^\alg \gg\pxa^\alg$. To mitigate the inequality, the decision maker can introduce selection quotas. One example is the $\ff$-rule \cite{holzer00} that imposes that $\pxa \ge \frac45 \pxb$ and $\pxb \ge \frac45\pxa$.

In this paper, we consider a generalization of the $\ff$-rule  that is parameterized by $\gamma \in [0, 1]$. We say that a selection satisfies the \emph{$\gamma$-rule} if
\begin{align}
  \pxa \ge \gamma \pxb \qquad\text{ and }\qquad
  \pxb \ge \gamma \pxa.
  \label{eq:beta-fair}
\end{align}
A selection algorithm satisfies this constraint if and only if it picks at least $m\gamma n_A/(n_B+\gamma n_A)$ $A$-candidates and at least $m\gamma n_B/(n_A+\gamma n_B)$ $B$-candidates. Indeed, the total number of selected candidates is $m=x_An_A+x_Bn_B$ which means that $\pxb=(m-\pxa n_A)/n_B$. The constraint $\pxa \ge \gamma \pxb$ is therefore true if $\pxa\ge\gamma x_B=\gamma(m-\pxa n_A)/n_B$ which is true if and only if $\pxa\ge\gamma m/(n_B+\gamma n_A)$.  Similarly, the constraint $\pxb \ge \gamma \pxa$ is true if and only if $\pxb\ge\gamma(m-\pxb n_B)/n_A$. 

This means that one can easily transform a baseline into a $\gamma$-fair algorithm by first selecting at least $m\gamma n_A/(n_B+\gamma n_A)$ $A$-candidates and at least $m\gamma n_B/(n_A+\gamma n_B)$ $B$-candidates and then filling the remaining positions according the best estimated candidates (candidates with largest $\whi$ if the baseline algorithm is group oblivious and  with largest $\wti$ if the baseline algorithm is Bayesian-optimal), irrespective of their group. This is what defines the \emph{$\gamma$-fair group-oblivious} and \emph{$\gamma$-fair Bayesian-optimal} algorithms.

When $\gamma=0$, the $\gamma$-fair version of a baseline algorithm reduces to the original unconstrained  algorithm (the algorithm that does  not to take into  account fairness). When $\gamma=1$, the $\gamma$-rule mechanism corresponds to the classical notion of \emph{demographic parity} \cite{Zafar17a} that mandates that the selection rates be equal across different groups. We highlight the demographic parity mechanism as a special and important case of the $\gamma$-rule. Note that because $n_A$, $n_B$ and $m$ are integer variables, it might be impossible to satisfy the constraints in~\eqref{eq:beta-fair} when $\gamma$ is too close to $1$. In such a case, we say that an algorithm is $\gamma$-fair if the constraint \eqref{eq:beta-fair} is satisfied up to one candidate.

\subsection{Simplification of the selection problem for large $n$ and $m$}

In the remainder of the paper, we study the selection problem when the
number of candidates is large.  That is, we assume that there exist fixed
fractions $\pa,\ax\in(0,1)$ such that 
\begin{align*}
  n_A = \floor{\pa n}\qquad n_B = \ceil{(1-\pa)n} \qquad m = \floor{\ax n},
\end{align*}
and let $n$ grow.  Our theoretical results are obtained in the limit where $n$ goes to
infinity (similarly to \cite{kleinberg18,celis20}). In Section~\ref{ssec:approximation} we  show numerically that our results for $n=\infty$ continue to hold for
finite selection sizes. Note that $p_A$ represents the fraction of $A$-candidates in the population while $\ax$ represents the global selection ratio (or budget).

For a finite $n$, the selection algorithms presented in Sections~\ref{sec:selection_algorithms}-\ref{ssec:fairness mechanisms} are hard to analyze because the probability for a candidate to be selected depends on all other candidates. As we prove below, characterizing the performance of a selection problem is simpler when the number of candidates $n$ is infinite because there is an equivalence between the algorithms presented in the previous sections and threshold-based algorithm. A threshold-based algorithm uses two thresholds $\twha$ and $\twhb$ and selects all $G_i$-candidates, such that $\wh_i \ge \twhgi$.\footnote{Note that the Bayesian-optimal algorithm can also be written that way, with appropriate thresholds, because \emph{within a given group} the expected qualities $\wti$ are in the same order as the signals $\whi$.} For given thresholds $\twha$ and $\twhb$, we denote the expected utility of the corresponding selection by $\V(\twha, \twhb)$:
\begin{align*}
  \V(\twha, \twhb) = %
  \esp{\w_i\,|\, \wh_i \geq \twhgi}.
\end{align*}
Hence, the selection of a candidate does not depend on the qualities of the other individuals. Also, as we show in the next theorem, the fraction of $A$-candidates that are selected becomes deterministic as $n$ goes to infinity. 
\begin{lemma}
  \label{th:n_infinity}
  For any of the selection algorithms presented in Sections~\ref{sec:selection_algorithms}-\ref{ssec:fairness mechanisms},
  \begin{enumerate}
  \item there exists a deterministic fraction $x_A\in[0,1]$ such that
    the fraction of $A$-candidates that are selected by the algorithm
    converges (in probability) to $x_A$ as $n$ grows;
  \item there exist deterministic thresholds $\twha, \twhb$ such
    that the expected utility of this algorithm converges to
    $\V(\twha, \twhb)$.
  \end{enumerate}
\end{lemma}
\begin{proof}[Proof Sketch]
  The above result is essentially a direct consequence of the law of
  large numbers.  By the Glivenko-Cantelli theorem, the empirical
  distribution of the estimated qualities of the $G$-candidates
  converges to the distribution of $\whg$ as $n\to\infty$. This shows
  that taking the best $\floor{n\pa\pxa}$ $A$-candidates or taking all
  $A$-candidates above the $\pxa$-quantile of the distribution $\wha$
  is asymptotically equivalent as $n\to\infty$. 
\end{proof}

For these given thresholds $\twha, \twhb$, the fractions of
selected candidates are $\Pb(\wh_i\ge\twhgi)$.  Using
the above definition, we denote by $\Q(\pxa)$ the expected utility of
a threshold-type selection algorithm that selects $A$-candidates with
probability $\pxa$ and that satisfies the selection size constraints in
expectation:
\begin{align}
  \label{eq: u}
  \Q(\pxa) &=\V(\twha, \twhb),\text{ where $\twha, \twhb$
             are such that}
             \left\{\begin{array}{l}
                      \Pb(\whi \ge \twha\,|\,G_i=A) = \pxa,\\
                      \Pb(\whi \ge \twhgi) = \ax.
                    \end{array}
  \right.
\end{align}
Note that combining the constraints in \eqref{eq: u} immediately gives that such an algorithm selects $B$-candidates with probability $\pxb = (\ax - \pxa \pa)/(1-\pa)$. Hence, it is sufficient to describe the algorithm with $\pxa$.


The above definition of expected quality is not directly applicable to
the selection algorithms presented in
Section~\ref{sec:selection_algorithms} because those algorithms are
defined neither in terms of fraction of selected candidates nor in
terms of thresholds. In fact, for a given selection algorithm, the
fractions of selected $A$- and $B$-candidates depend on the
realizations of the random variables representing the quality ($\wi$)
and the estimated quality ($\whi$). As a result, these fractions ($\pxa$
and $\pxb$) are random variables. For instance, if because of
randomness the $A$-candidates are evaluated much worse than
the $B$-candidates, then $\pxa$ will be $0$ for the group-oblivious algorithm.
Lemma~\ref{th:n_infinity} shows that when the population is large, these random fluctuations disappear. It shows that, when $n$ is large, the performance of the various algorithms are simply characterized by $\pxa$.



For a finite $n$, characterizing precisely the utility of an algorithm like group-oblivious is computationally difficult due to the correlations between the selection of the different agents. Lemma~\ref{th:n_infinity} allows us to greatly simplify the
study of the performance of the various algorithms because the function $\Q$, defined in \eqref{eq: u}, depends only on one parameter $\pxa$, and is simpler to characterize than the expectation over a finite number of candidates $n$.

\subsection{Summary of main notation}

We denote respectively by  $\pxagreedy$, $\pxafair$,
$\pxaopt$ and $\pxafopt$  the asymptotic fraction of $A$-candidates that
are selected for the group-oblivious, the $\gamma$-fair group-oblivious,
 the Bayesian-optimal and the $\gamma$-fair Bayesian-optimal algorithms. We also identify an important subcase of the $\gamma$-rule for $\gamma=1$. In this case both the $\gamma$-fair group-oblivious algorithm and the $\gamma$-fair Bayesian-optimal algorithm select $A$-candidates at rate $\pxa=\ax$, so there is no difference between them. We name the corresponding selection algorithm as \emph{demographic parity algorithm}. 

We denote the expected performance of the introduced algorithms by
\begin{align*}
  \Qgreedy = \Q(\pxagreedy); %
  \qquad \Qfair = \Q(\pxafair); %
  \qquad \Qopt = \Q(\pxaopt); %
  \qquad \Qfopt = \Q(\pxafopt); %
  \qquad \Qdp = \Q(\pxadp). %
\end{align*}
We summarize the other notation in Table~\ref{table: notation}.
\begin{table}[ht]
  \centering
  \caption{Summary of notation.}
  \label{table: notation}
  \vspace{-.2cm}  \begin{tabular}{ll}
    \toprule
    $\wi$ & latent quality of candidate $i$\\
    $\whi$ & estimated quality of candidate $i$\\
    $\wti$ & expected value of latent quality of candidate $i$ given the estimate $\whi$\\
    \midrule
    $\mqg$ & expected value of latent quality $\wg$\\
    $\sqg^2$ & variance of latent quality $\wg$\\
    $\sxg^2$ & variance of additive noise\\
    $\ssg^2$ & variance of estimated quality $\whg$. It equals $\sxg^2+\sqg^2$.\\
    $\stg^2$ & variance of expected quality $\wtg$. It equals $\sqg^4 / \left(\sqg^2 + \sxg^2\right)$\\
    \midrule
    $\pxg^\alg$ & fraction of $G$-candidates that are selected by a given algorithm ``$\alg$''\\
    $\twhg^\alg$ & threshold above which $G$-candidates are selected by the algorithm ``$\alg$''\\
    \midrule
    $\phi$, $\Phi$, $\Phi^{-1}$ & PDF, CDF and quantile of the standard normal distribution $\N(0,1)$\\
    \bottomrule
\end{tabular}
\end{table}

\section{Analysis of the general model}
\label{section: one-stage general case}

In this section, we present the main technical results of the paper in the most general model. The results that we prove in this section are quite abstract; to make things more concrete and provide more intuitive results, we will instantiate this general model in important sub-cases in Section~\ref{section: cases}.

We start by showing why the two baseline algorithms lead to discrimination, in Theorem~\ref{theorem: group oblivious general case} for the group-oblivious and in Theorem~\ref{theorem: bayesian optimal general case} for the Bayesian-optimal algorithm. Then, we specify in Theorem~\ref{theorem: group dependent prior} conditions under which the $\gamma$-rule fairness mechanism  increases the utility of selection compared to the unconstrained group-oblivious algorithm. Although it is clear that the $\gamma$-rule mechanism cannot increase the utility of the Bayesian-optimal algorithm (since it is an expected utility and the Bayesian-optimal algorithm maximizes it by definition), we prove in Theorem~\ref{theorem:cost of fairness general} that the ratio of the utilities of the unconstrained Bayesian-optimal and the $\gamma$-fair Bayesian-optimal algorithms is bounded.

\subsection{Discrimination of baseline selection algorithms}
\label{ssec:one-stage:policies}

Recall that we assume (without loss of generality) that group $A$ is the high-variance group, that is $\ssa^2>\ssb^2$. Then, the distribution of $\wha$ has longer tails compared to the distribution of $\whb$.  Thus, if the selection size is small, $A$-candidates will be selected by the group-oblivious algorithm at higher rate compared to $B$-candidates because the probability to estimate an $A$-candidate as ``outstanding'' is higher than for $B$-candidates. In contrast, if the selection size is large, the chance of estimating an $A$-candidate as poor is larger than for $B$-candidates, in which case the group-oblivious algorithm selects a lower fraction of $A$-candidates. This can be formally stated as follows.
\begin{theorem}
  \label{theorem: group oblivious general case}
  Assume without loss of generality that $\ssa^2 > \ssb^2$. When using the group-oblivious selection algorithm, the selection rates for
  $A$- and $B$-candidates, $\pxagreedy{}$ and $\pxbgreedy{}$, satisfy:
  \begin{align*}
    \pxagreedy{} > \pxbgreedy{} \text{ if and only if } \ax < \Phi\left(\frac{\dmq - \db}{\dsx}\right),
  \end{align*}
  where $\dmq=\mqa-\mqb$, $\db=\ba-\bb$ and $\dsx = \ssa - \ssb$.
\end{theorem}

\begin{proof}[Proof Sketch]
The group-oblivious algorithm sorts candidates by their estimated qualities $\whi$ and takes the best $\ax n$ by applying a group-independent threshold $\tx$. The expression for the selection rates $\pxggreedy{}=1- \Phi\left(\frac{\tx - \mqg + \bg}{\ssg}\right)$ and a simple rearrangement allows us to find such sizes of budget $\ax$ for which  selection rates for both groups become equal $\pxagreedy{}=\pxbgreedy{}$. The result then follows from the corresponding properties of normal CDF and our assumption that $\ssa > \ssb$. A detailed proof is given in \ref{proof:group oblivious general case}.
\end{proof}

The above result implies that, for a small selection budget, the group-oblivious algorithm will select high-variance candidates at a higher rate. Note that this result does not assume that this higher variance comes from the variance of the true quality ($\sqa^2$ and $\sqb^2$) or from the variance of the estimates ($\sxa^2$ and $\sxb^2$). It is only assumed that the variance of $\wha$, equal to $\ssa^2=\sxa^2 + \sqa^2$, is larger than the one of $\whb$.

As we show below, nearly the opposite is true for the Bayesian-optimal algorithm: for a small selection budget, in the case of group-independent variance of the latent quality ($\sqa^2=\sqb^2$), a Bayesian-optimal algorithm will select fewer candidates from the high-variance group.  In the case where $\sqa^2\not=\sqb^2$, though, which group is underrepresented will be determined by the variances $\stgeneric$ and not $\ssgeneric$, see our discussion below the theorem. Note also that the specific budget threshold at which the transition happens is not the same as for the group-oblivious algorithm.
\begin{theorem}
  \label{theorem: bayesian optimal general case}
  Assume that $\sta^2 < \stb^2$. When using the Bayesian-optimal selection algorithm, the selection rates for $A$- and $B$-candidates, $\pxaopt$ and $\pxbopt$, satisfy:
  \begin{align*}
    \pxaopt < \pxbopt{} \text{ if and only if } \ax < \Phi\left(\frac{\dmq}{\dst}\right),
  \end{align*}
  where $\dmq = \mqa - \mqb$, $\dst = \sta - \stb$.
  \end{theorem}

\begin{proof}[Proof Sketch]
In the Bayesian-optimal algorithm, the candidates are sorted by their expected qualities $\wt$ and a group-independent threshold is applied to select the best $\ax n$ candidates. The expression \eqref{eq:expected w given w hat} for the expected quality $\wtg$ allows us to compare the selection rates $\pxaopt$ and $\pxbopt$ for different groups $A$ or $B$ and to find such value of budget $\ax$ for which $\pxaopt=\pxbopt$. Then using the fact that $\wtg$ follows normal law and the relation between $\sta$ and $\stb$, we obtain our result. A complete proof can be found in \ref{proof:bayesian optimal general case}.
\end{proof}

The result of Theorem~\ref{theorem: bayesian optimal general case} is consistent with the observation from Phelps  \cite{phelps72} in a simpler setting (without bias and with group-independent distribution of the latent quality $\w$): in the presence of differential variance, the candidates from the high-variance group will appear more similar to each other to the decision maker, hence the distribution of computed expected quality will have a longer tail for the low-variance group. As a consequence, for small enough selection budgets, candidates from the high-variance group will be selected at a lower rate.

Note that the result in Theorem~\ref{theorem: bayesian optimal general case} imposes a condition on the order between $\sta^2=\sqa^4 / \left(\sqa^2 + \sxa^2\right)$, the variance of $\wta$, and $\stb^2$; but it is not conditional on the relation between $\ssa^2$ and $\ssb^2$, i.e., both $\ssa^2 > \ssb^2$ and $\ssa^2 \le \ssb^2$ are allowed. Hence the condition $\sta^2 < \stb^2$ comes without loss of generality for this result. 
Note also that in the case where the variances of the true quality are the same for both groups ($\sqa=\sqb$), the two conditions from Theorems~\ref{theorem: group oblivious general case} and \ref{theorem: bayesian optimal general case} are equivalent, that is $\sta<\stb$ if and only if $\ssa > \ssb$ (since it holds if and only if $\sxa>\sxb$). The main special cases that we consider in Section~\ref{section: cases} (specifically those of Sections~\ref{ssec:group-independent quality} and \ref{ssec:bias}) are in this case (i.e., satisfy $\sqa=\sqb$).

\subsection{The $\gamma$-rule mechanism can increase the utility of the group-oblivious algorithm}
\label{ssec:improve}

As we  show in Theorem~\ref{theorem: group oblivious general case}, the group-oblivious algorithm leads to overrepresentation of the high-variance group $A$, if the budget $\ax$ is small. To mitigate this effect, the decision maker can use the $\gamma$-rule fairness mechanism introduced in  Section~\ref{ssec:fairness mechanisms}. 

In the next theorem, we provide a condition on budgets $\ax$ for which using the $\gamma$-fair group-oblivious algorithm attains larger quality of selection compared to the unconstrained group-oblivious algorithm. The main message of this theorem is that if  $A$-candidates have larger variability of their estimate compared to $B$-candidates (i.e., $\ssa^2 > \ssb^2$) and the variance of expected quality for $A$-candidates is smaller than for $B$-candidates (i.e., $\sta^2 < \stb^2$), then the $\gamma$-fair group-oblivious algorithm leads to larger quality of selection compared to the group-oblivious algorithm for both small and large  budgets $\ax$.  As said earlier, when $\sqa=\sqb$, these conditions are always satisfied, up to switching the groups $A$ and $B$. At the same time, there may exist a region of budgets $\ax$ such that the $\gamma$-rule fairness mechanism harms the quality of selection compared to the group-oblivious algorithm.
\begin{theorem}
\label{theorem: group dependent prior}
Without loss of generality, assume that the estimates of quality for $A$-candidates  has larger variance than for $B$-candidates $\ssa^2 > \ssb^2$. Assume also that the variance of the expected quality is smaller for $A$-candidates than for $B$-candidates ($\sta^2 < \stb^2$), and let us define \begin{align*}
  \amin=\min\left\{\Phi\left(\frac{\dmq - \db}{\dsx}\right), \Phi\left(\frac{\dmq}{\dst}\right)\right\},\; 
  \amax=\max\left\{\Phi\left(\frac{\dmq - \db}{\dsx}\right), \Phi\left(\frac{\dmq}{\dst}\right)\right\},
\end{align*} where $\dmq=\mqa-\mqb$, $\db=\beta_A-\beta_B$, $\dsx = \ssa-\ssb$ and $\dst = \sta-\stb$. We have:
\begin{enumerate}[(i)]
  \item For any $\ax \in (0, \amin) \cup (\amax, 1)$,
 the demographic parity algorithm strictly improves the selection quality compare to the group-oblivious algorithm and the $\gamma$-fair group-oblivious algorithm for $\gamma< 1$ weakly improves it:
$$\Qdp > \Qfair \ge \Qgreedy{}.$$
\item If $\amin=\amax$, then for $\ax=\amin=\amax$ one has $\Qdp = \Qfair = \Qgreedy{}$.
\item Assume that $\amin\ne\amax$, then there exists $[\tilde \ax^{\mathrm{min}}, \tilde \ax^{\mathrm{max}}]$, where $\tilde\ax^{\mathrm{min}} >  \amin$ and  $\tilde \ax^{\mathrm{max}} < \amax$, such that for any $\ax \in [\tilde \ax^{\mathrm{min}}, \tilde \ax^{\mathrm{max}}]$, the demographic parity algorithm strictly harms the selection quality compared to the group-oblivious algorithm and the $\gamma$-fair group-oblivious algorithm for $\gamma< 1$ weakly harms it:
 $$\Qdp < \Qfair \le \Qgreedy{}.$$
\end{enumerate}
\end{theorem}

\begin{proof}
    We prove in Theorem~\ref{theorem: group oblivious general case} that if $\ax  < \Phi\left((\dmq - \db)/\dsx\right)$, then the group-oblivious algorithm leads  to overrepresentation of the high-variance group $A$. At the same time, the group-oblivious algorithm leads to underrepresentation of the group $A$ if  $\ax > \Phi\left((\dmq - \db) / \dsx\right)$. Similarly, we prove in Theorem~\ref{theorem: bayesian optimal general case} that if $\sta < \stb$, then for $\ax  < \Phi\left(\dmq/\dst\right)$, the Bayesian-optimal algorithm underrepresents the group $A$ and for $\ax > \Phi\left(\dmq/\dst\right)$  it overrepresents the group $A$.

    Recall that for any value of $\ax$, the demographic parity algorithm requires that candidates from both groups, $A$ and $B$, must be selected at equal rates, i.e.,  $\pxadp=\pxbdp=\ax$. It means  that if $\ax \in (0, \amin)\cup (\amax, 1)$, then the demographic parity algorithm will  perform a selection such that either $\pxagreedy{} < \pxadp < \pxaopt$ or $\pxaopt < \pxadp < \pxagreedy$ (also $\pxagreedy{} \le \pxafair 
    < \pxaopt$ or $\pxaopt < \pxafair 
    \le \pxagreedy$ for $\gamma < 1$). In \ref{ssec: properties of utility}, we prove that the selection quality $\Q$ is a concave function of $\pxa$ with a single maximum at $\pxa=\pxaopt$. Hence, from this property we conclude that $\Qdp > \Qgreedy{}$ and $\Qfair \ge \Qgreedy{}$ for $\gamma < 1$. Finally, (iii) is due to  the fact that the utility $\Q$ is a continuous and smooth function of  $\pxa$  as we prove in \ref{ssec: properties of utility}.
\end{proof}

While the statement of Theorem~\ref{theorem: group dependent prior} is somewhat complex due to its generality, in special cases (e.g., that of Section~\ref{ssec:group-independent quality}) we have $\amin=\amax$. This means that in the special case of Section~\ref{ssec:group-independent quality}, we are always in case (i) of Theorem~\ref{theorem: group dependent prior}: the $\gamma$-fair group-oblivious algorithm attains a larger utility than the corresponding baseline (or at worst an equal utility).


Note that the statement of Theorem~\ref{theorem: group dependent prior} is under the assumption that $\ssa^2>\ssb^2$ \emph{and} $\sta^2 < \stb^2$. As discussed earlier, this assumption may not be without loss of generality if $\sqa\neq\sqb$. 
If it does not hold, then using the demographic parity algorithm could lead to a worse utility than the group-oblivious algorithm. Even in this case, however, the ratio $\Qgreedy{}/\Qdp$ remains bounded. Indeed, we can write $\Qgreedy{}/\Qdp \le \Qopt/\Qdp$ and the ratio $\Qopt/\Qdp$ itself is upper-bounded as we show in the next section (Theorem~\ref{theorem:cost of fairness general}).

\subsection{Bounds on the decrease of utility due to imposing $\gamma$-rule on the Bayesian-optimal algorithm}
\label{ssec:harm}

By definition, the Bayesian-optimal algorithm maximizes the utility of the selection which means that imposing a $\gamma$-rule cannot increase the expected utility of the selection---in most cases it decreases it. In this section, however, we obtain a bound on the ratio of utilities for Bayesian-optimal and $\gamma$-fair Bayesian-optimal algorithms. This is stated in the following theorem:

\begin{theorem}
\label{theorem:cost of fairness general}
  Assume that $\sta^2 < \stb^2$ and that $\mqa,\mqb\ge0$, then for any budget $\ax$ the ratio $\Qopt{} / \Qfopt$ satisfies the following bound:
  \begin{align*}{}
    1 \le \frac{\Qopt{}}{\Qfopt{}} \le 1 + 
  \begin{cases} 
     - \frac{\ax}{\pa + \pb/\gamma}\cdot g(\mqg, \stg, \pg, \ax),  & \text{ if } \ax \le \Phi\left(\frac{\dmq}{\dst}\right)\\
    \left(1 - \frac{\ax}{\pa + \pb\gamma}\right)\cdot g(\mqg, \stg, \pg, \ax),   & \text{ if } \ax > \Phi\left(\frac{\dmq}{\dst}\right)
  \end{cases}
\end{align*}
   where $\dmq = \mqa - \mqb$, $\dst = \sta- \stb$ and $g(\mqg, \stg, \pg, \ax) = \frac \pa \ax\frac{\dmq + \Phi^{-1}(1-\ax)\dst}{\sum_G\pg\mqg + \frac{\phi\left(\Phi^{-1}(1-\ax)\right)}{\ax}\sum_G \pg\stg}$. 
\end{theorem}

\begin{proof}[Proof Sketch]
  The first inequality is due to the fact that the utility function $\Q(\pxa)$ is strictly concave and that it attains its maximum at $\pxa=\pxaopt$ as we show in \ref{ssec: properties of utility}.

  To prove the second inequality, we need a few preparatory steps.
  First, using the result of Theorem~\ref{theorem: bayesian optimal general case}, we obtain that for the budgets $\ax < \Phi\left(\dmq/\dst\right)$, we have $\pxaopt{} \le \pxafopt < \pxadp$. Using the concavity of $\Q$ and the mean value theorem from real analysis, we obtain:
    $\frac{\Q(\pxaopt) - \Q(\pxadp)}{\pxaopt - \pxadp} \ge \Q'(\pxa=\pxadp) \implies \Q(\pxaopt) - \Q(\pxadp) \le -\ax \cdot \Q'(\pxadp).$
    After, we divide both parts by $\Q(\pxa=\pxadp)$.
    The expressions for $\Q'(\pxa=\pxadp)$ and $\Q(\pxa=\pxadp)$ can be written explicitly using the equation derived in \ref{ssec: properties of utility}. A complete proof is given in \ref{proof:cost of fairness}.
  \end{proof}

The expression in Theorem~\ref{theorem:cost of fairness general} is general but complex due to the large number of model parameters. It can be simplified as we tighten up some of the assumptions (see Section~\ref{section: cases}). There are also interesting behaviors to observe for some values of the parameters. First, if the size of group $A$ becomes small (i.e., $\pa \to 0$), we observe that the function $g$ converges to 0, hence the upper bound converges to 1. This is expected, since the introduction of the $\gamma$-rule mechanism will affect the selection in a tiny amount due to a small number of $A$-candidates. Second, as the selection budget decreases (i.e., $\ax \to 0$), we can show using L'H\^opital's rule that the upper bound in this limit converges to $1 - \frac{\pa \dst}{\sum \pg \stg}$ for $\gamma=1$. In other words, the difference in the expected values of qualities $\dmq$ does not play any role. This is also quite natural, since for tiny selection budgets $\ax$, the competition is among the candidates with very large values of quality which is due to the variance of the distribution of latent quality but not their mean values.

\section{Notable special cases of the general model}
\label{section: cases}


The results in Section~\ref{section: one-stage general case} might be difficult to interpret without considering some specific cases. In this section, we decompose  the effects of different factors by tightening up some of the assumptions of our model while keeping the others in place. We consider the following important special cases: In Section~\ref{ssec:group-independent quality} we assume that there is no bias in the estimation of quality and that the quality distribution is group-independent.  This is the model studied in \cite{emelianov20}, where the only quantity that depends on the candidate's group is the noise variance (to isolate the differential variance effect). In Section~\ref{ssec:bias}, we assume that the quality distribution is  group-independent but the estimates are biased.  In Section~\ref{ssec:group-dependent} we assume unbiased estimates but let the quality be group-dependent. All these subcases allow us to greatly simplify the results of Theorem~\ref{theorem: group dependent prior} and Theorem \ref{theorem:cost of fairness general}.

\subsection{Group-independent latent quality and unbiased estimates}
\label{ssec:group-independent quality}

In this section, we assume that the underlying quality distribution is group-independent (this is the classical assumption in the literature, see for instance \cite{kleinberg18,celis20}) and  follows a normal law with mean $\mq$ and variance $\sq^2$. To isolate the effect of the variance, we also assume that quality estimates $\whi$ are unbiased, i.e., $\bgi=0$. The main result of this section is that imposing a fairness constraint in this context \emph{cannot} decrease the utility compared to using the unconstrained group-oblivious baseline. 
We also simplify the bound of Theorem~\ref{theorem:cost of fairness general} on the decrease of the utility of the Bayesian-optimal algorithm due to the $\gamma$-rule fairness mechanism.

First, the following corollary relates selection ratios for two baseline algorithms. It can be obtained directly from Theorems~\ref{theorem: group oblivious general case} and \ref{theorem: bayesian optimal general case}. (Recall that in this special case of group-independent quality distribution, we have $\sxa^2 > \sxb^2$ if and only if $\ssa^2 > \ssb^2$, which is also if and only if $\sta^2 < \stb^2$.)
\begin{corollary}[Corollary of Theorems~\ref{theorem: group oblivious general case} and \ref{theorem: bayesian optimal general case}]
  \label{corollary: selection for go}
  Assume that the quality distribution is group-independent $\wi \sim \N(\mq,\sq^2)$ and that the quality estimates $\whi$ are unbiased $\bg=0$, $\forall G \in \{A,B\}$.  Assume without loss of generality that $\ssa^2 > \ssb^2$. When using the group-oblivious selection algorithm and the Bayesian-optimal selection algorithm, the fractions $\pxggreedy$ and $\pxgopt$ of selected candidates from each group satisfy:
  \begin{enumerate}[(i)]
  \item $\pxagreedy > \pxbgreedy$ if and only if $\ax < 1/2$;
  \item $\pxaopt < \pxbopt$  if and only if $\ax < 1/2$.
  \end{enumerate}
\end{corollary}

Corollary~\ref{corollary: selection for go} formalizes in simple terms, for the case of group-independent latent quality distributions and unbiased estimators, the discrimination that results from the two baseline algorithms. Notably, (i) states that for selection budgets below $1/2$, the group-oblivious algorithm overrepresents the high-variance group. If the high-variance group is a minority, this is counter-intuitive. As noted in the introduction, however, these typically correspond to cases where the Bayesian-optimal baseline is more meaningful. Then, (ii) states that for small budgets, the Bayesian-optimal algorithm indeed underrepresents the high-variance group.

In Section~\ref{section: one-stage general case} we specify a condition under which the $\gamma$-rule fairness mechanism is beneficial to the utility of the group-oblivious algorithm.  In the special case of group-independent prior and unbiased estimator, the thresholds $\amin$ and $\amax$ defined in Theorem~\ref{theorem: group dependent prior} coincide and are equal to $1/2$. This implies the next theorem, which shows that for this case, the $\gamma$-rule fairness mechanism cannot decrease the average quality of a selection compared to the group-oblivious algorithm (without any condition on $\ax$).
\begin{corollary}[Corollary of Theorem~\ref{theorem: group dependent prior}]
  \label{theorem: 1 stage}
  Assume that the quality distribution is group-independent $\wi \sim \N(\mq,\sq^2)$ and that the quality estimates $\whi$ are unbiased $\bg = 0$, $\forall G \in \{A,B\}$.  Let, without loss generality, $\ssa^2 > \ssb^2$, then
  for any budget $\ax\ne1/2$, the
  demographic 
  parity selection algorithm provides a larger utility than the $\gamma$-fair group-oblivious selection algorithm with $\gamma<1$, which in
  turn provides a larger utility than the group-oblivious selection algorithm:
  \begin{align*}
    \Qdp>\Qfair\ge\Qgreedy.
  \end{align*}
  The above inequality is an equality when $\ax=1/2$.
\end{corollary}

\begin{proof}
  This result is  a special case of Theorem~\ref{theorem: group dependent prior}. Since  the distribution of quality is group-independent and there is no implicit bias, then the condition in Theorem~\ref{theorem: group dependent prior} holds, and  $\amin$ and $\amax$ coincide and become equal to $1/2$.
\end{proof}


As for the general case, the $\gamma$-rule fairness mechanism cannot increase the selection quality of the Bayesian-optimal baseline. In  Theorem~\ref{theorem:cost of fairness general}, we obtained a bound on the decrease of utility. In the next result, we show how this result simplifies in the modeling assumptions of the current subsection. We provide the bound for $\ax \le 1/2$ as it is the most interesting setting. The one for $\ax > 1/2$ can also be easily deduced.
\begin{corollary}[Corollary of Theorem~\ref{theorem:cost of fairness general}]
  \label{theorem:cost of fairness}
  Assume that quality distribution is group-independent $\wi \sim \N(\mq,\sq^2)$ and quality estimates $\whi$ are unbiased $\bg = 0$, $\forall G \in \{A,B\}$.
  Let, without loss of generality, $\ssa^2 > \ssb^2$ and $\mq\ge 0$. Then for all $\ax \not =  1/2$, the
  demographic 
  parity selection algorithm provides a smaller utility than the $\gamma$-fair Bayesian-optimal selection algorithm with $\gamma<1$, which in
  turns provides a smaller utility than the Bayesian-optimal selection algorithm. The utility ratio $\Qopt/\Qfopt$ for any budget $\ax \le 1/2$ has the following bound:
  $$1 \le \frac{\Qopt{}}{\Qfopt{}} \le 1 + g(\ax)\cdot\frac{1}{\pa + (1-\pa)/\gamma}\cdot\frac{\pa(\nu - 1)}{\pa + (1-\pa)\nu},$$
  where $g(\ax) = \frac{\ax \Phi^{-1}(1-\ax)}{\phi\left(\Phi^{-1}\left(1-\ax\right) \right)}$ and $\nu = \ssa/\ssb > 1$. 
\end{corollary}

\begin{proof}
  Direct from Theorem~\ref{theorem:cost of fairness general} when we set $\mq=\mqa=\mqb$ and $\sq^2=\sqa^2=\sqb^2$.
\end{proof}

For $\gamma=1$, which is the case of demographic parity, we can further simplify the expression in Corollary~\ref{theorem:cost of fairness}. By using the fact that $g(\ax)$ is decreasing with $\ax$ and that $\lim_{\ax \to 0} g(\ax) = 1$, we can write
   \begin{align*}
     1\le \Qopt / \Qdp \le 1 + \frac{\pa(\nu - 1)}{\pa + (1-\pa)\nu}.
   \end{align*} 
  Note that as $\nu$ tends to 1, meaning that there is no difference in variances between $A$ and $B$ group, the upper bound also tends to 1 and matches the lower bound.  Most interestingly, we observe that the larger the difference in variances $\nu$, the larger the upper bound. As $\nu$ tends to infinity, the upper bound tends to $1/(1-\pa)$. Hence, if, for instance, the high-variance group is the minority ($\pa < 1/2$), then the gap cannot be larger than 2.

\subsubsection*{Numerical illustrations} 
\begin{figure}
    \centering
    \begin{subfigure}{.24\textwidth}
      \includegraphics[width=\linewidth]{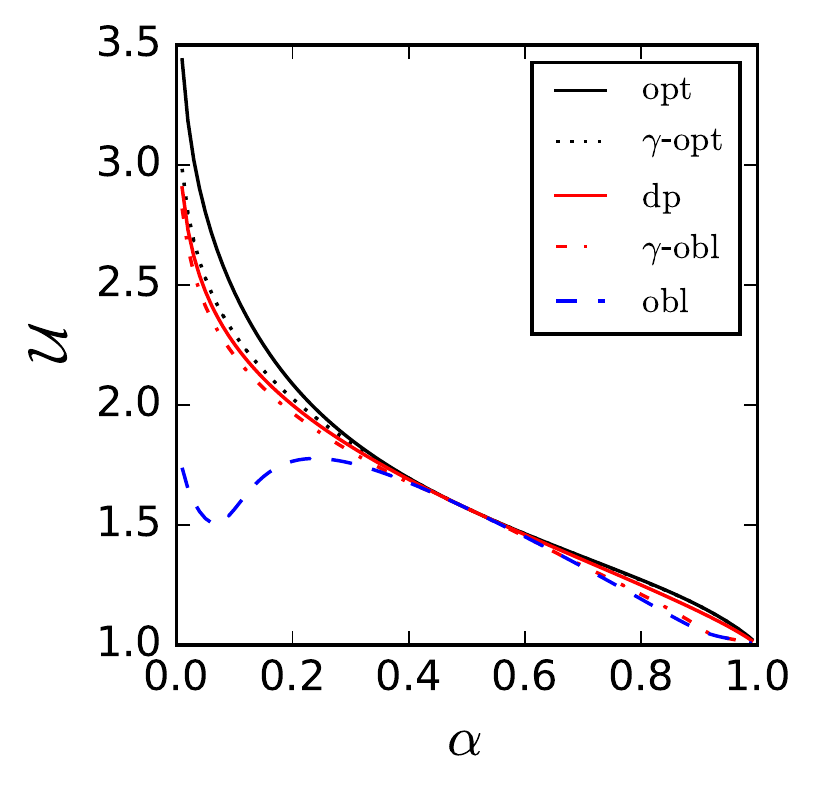}
      \vspace{-.6cm}\caption{Utility}
      \label{fig: all 1 stage_a}
    \end{subfigure}
    \begin{subfigure}{.24\textwidth}
      \includegraphics[width=\linewidth]{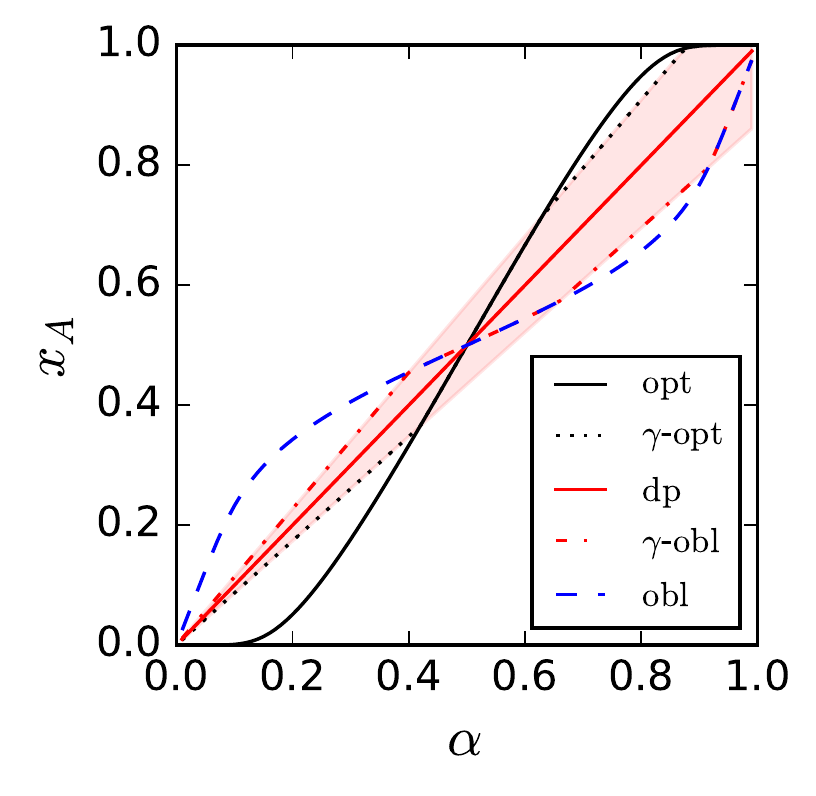}
      \vspace{-.6cm}\caption{Selection fraction}
      \label{fig: all 1 stage_b}
    \end{subfigure}
    \begin{subfigure}{.24\textwidth}
        \includegraphics[width=\linewidth]{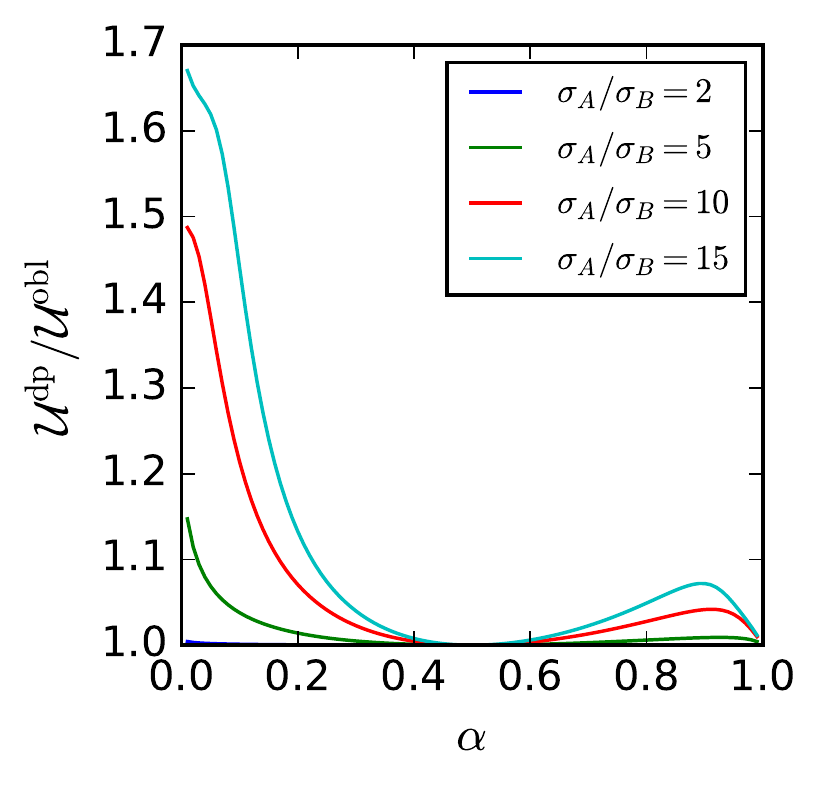}
      \vspace{-.6cm}\caption{Ratio $\Qdp/\Qgreedy{}$}
      \label{fig: all 1 stage_c}
    \end{subfigure}
    \begin{subfigure}{.24\textwidth}
      \includegraphics[width=\linewidth]{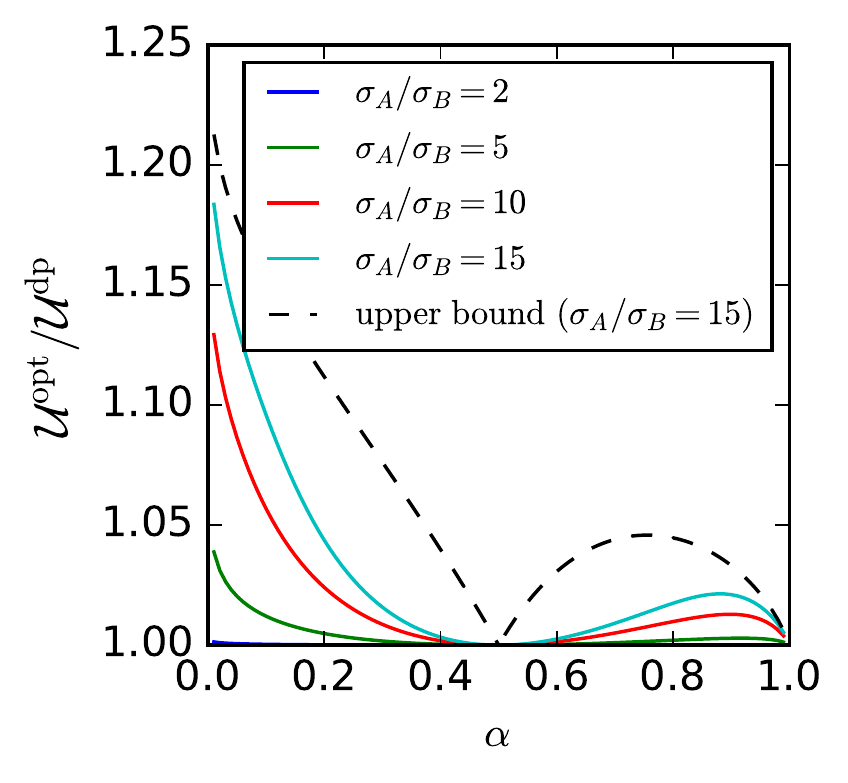}
    \vspace{-.6cm}\caption{Ratio $\Qopt/\Qdp$}
    \label{fig: all 1 stage_d}
  \end{subfigure}
    \vspace{-.2cm}\caption{Utility $\Q$, selection
      fraction $\pxa$ and performance gaps for
      different budgets $\ax$. The parameters are $\mq=1$,
      $\sq=1$, $\sxb=0.2$, and $\pa=0.4$; $\sxa=3$ for panels~(a,b). \vspace{-2mm}}
    \label{fig: all 1 stage}
\end{figure}

In Fig.~\ref{fig: all 1 stage}, we show the obtained utilities $\Q$, the selection fractions $\pxa$ and the gap values $\Qdp{}/\Qgreedy{}$ and $\Qopt/\Qdp$ for different budgets $\ax$ from 0.01 to 0.99.  Fig.~\ref{fig: all 1 stage_a} illustrates the utilities corresponding to different selection algorithms.  We observe that the utilities of the Bayesian-optimal and demographic parity selections decrease when $\ax$ increases. This is expected because this graph represents the average quality of a selected candidate: the average quality decreases when the number of selected candidates increases. What is more surprising is that the behavior of the group-oblivious selection algorithm is not monotonous: the expected utility $\Q$ increases when $\ax$ goes from $0.1$ to $0.3$. In fact, when $\ax<0.1$, very few $B$-candidates are selected by the group-oblivious algorithm. When $\ax\approx0.1$--$0.2$, this algorithm selects a few good $B$-candidates which leads to an increased average performance.

In Fig.~\ref{fig: all 1 stage_c} we show the performance gap between group-oblivious and demographic parity selection algorithms for different values of $\sxa$ and fixed $\sxb=0.2$, $\sq=1$. The values of $\sxa$ are such that $\sxa/\sxb=k$, $k=2,5,10,15$. We see that the gap is in general larger when the selection size $\ax$ is small. This is due to the fact that as the selection size increases, the selections by the group-oblivious and demographic parity algorithms become close. The performance gap is zero when $\ax = 0.5$ because the selections are exactly the same (due to the symmetry of the underlying quality distribution), but it becomes positive again for larger values of $\ax$. In addition, the larger the differential variance ratio $\sxa^2/ \sxb^2$, the larger the gain that demographic parity brings.

Finally, in Fig.~\ref{fig: all 1 stage_d} we illustrate the performance gap between the Bayesian-optimal and the demographic parity selection algorithms for different values of $\sxa$ and fixed $\sxb=0.2$, $\sq=1$. As in Fig.~\ref{fig: all 1 stage_c}, the values of $\sxa$ are such that $\sxa/\sxb=k$, $k=2,5,10,15$.
In addition, we also show the bound on  the ratio $\Qopt{}/\Qdp$ for different values of $\ax$ and for fixed $k=15$. We see that the upper bound developed in Theorem~\ref{theorem:cost of fairness general} is relatively tight for small values of $\ax$, but is quite loose when $\ax\approx0.5$.

\subsection{Group-independent latent quality distribution and biased estimates}
\label{ssec:bias}

In this section, we again assume that the underlying quality distribution is group-independent but we now assume that the estimates are both biased and with differential variance. 
Since  the true quality distribution is group-independent, then $\mq=\mqa=\mqb$ and $\sq^2=\sqa^2=\sqb^2$. Recall that in this case, the conditions in Theorem~\ref{theorem: group dependent prior} hold since under the assumption of  $\ssa^2 > \ssb^2$ which is w.l.o.g, the requirement  $\sta^2 < \stb^2$ is also satisfied.  The expressions for the budgets $\amin$  and $\amax$ specified in Theorem~\ref{theorem: group dependent prior} can also be simplified:
$$\amin=\min\left\{\Phi\left(\frac{-\db}{\dsx}\right), \frac 1 2\right\}\;, \amax=\max\left\{\Phi\left(\frac{-\db}{\dsx}\right), \frac 1 2\right\}.$$

\begin{figure}
    \centering
    \begin{subfigure}{.3\textwidth}
      \includegraphics[width=\linewidth]{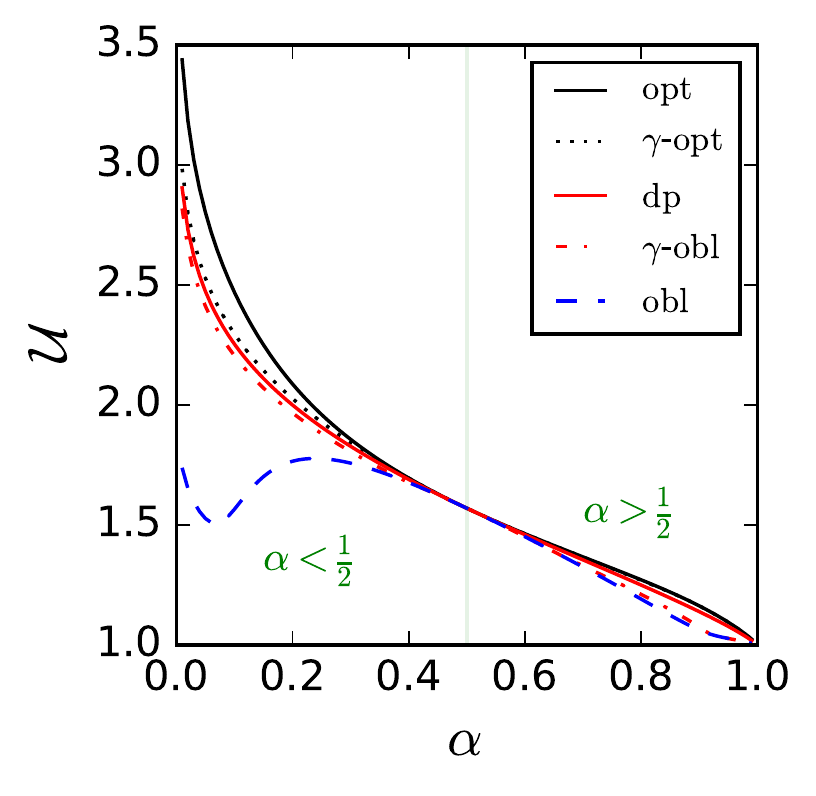}
      \vspace{-.5cm}\caption{$\ba = \bb$}
      \label{fig:bias_a}
    \end{subfigure}
    \begin{subfigure}{.3\textwidth}
      \includegraphics[width=\linewidth]{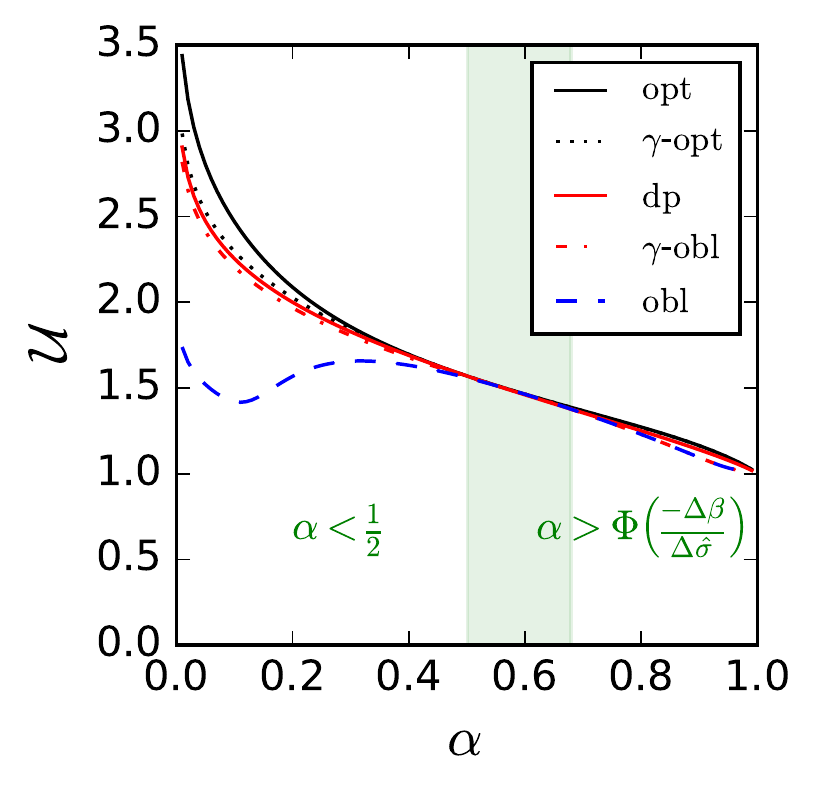}
      \vspace{-.5cm}\caption{$\ba < \bb$}
      \label{fig:bias_b}
    \end{subfigure}
    \begin{subfigure}{.3\textwidth}
      \includegraphics[width=\linewidth]{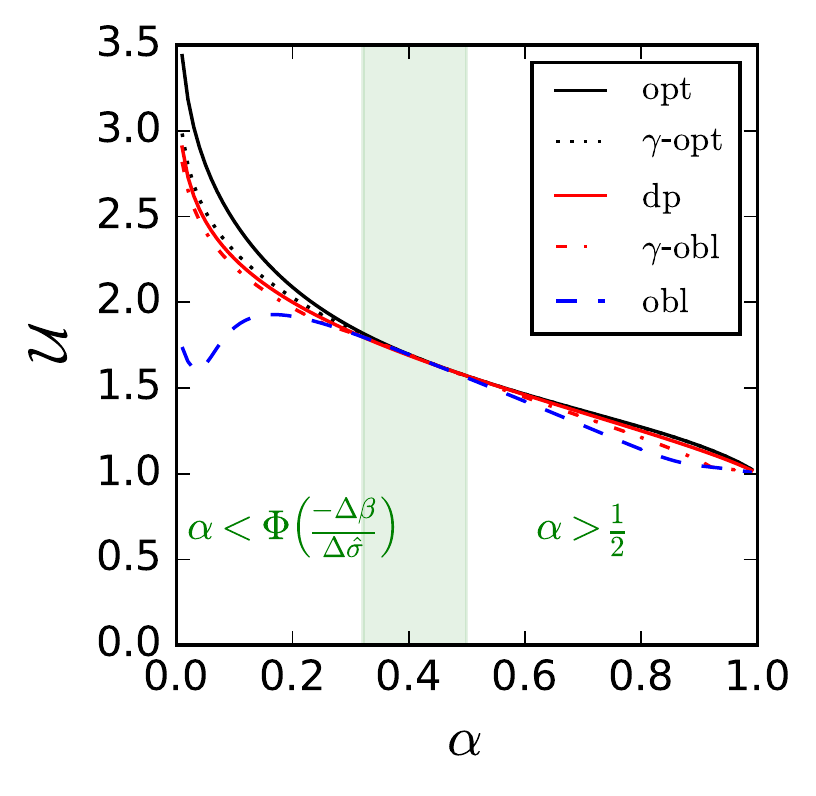} 
      \vspace{-.5cm}\caption{$\ba >\bb$}
      \label{fig:bias_c}
    \end{subfigure}
    \caption{The quality of selection in the presence of bias  and differential variance for different budgets $\ax$. We assume that the quality distribution is group-independent, but $A$-candidates have larger variability of estimation compare to the $B$-candidates, i.e. $\sxa^2 > \sxb^2$. The quality distribution follows $\N(\mq=1, \sq^2=1)$, the differential variance parameters are equal to $\sxa=3$ and $\sxb=0.2$. The bias parameters are equal  to $\ba=1,\bb=1$ for \ref{fig:bias_a}, $\ba=0,\bb=1$ for \ref{fig:bias_b} and $\ba=1,\bb=0$ for \ref{fig:bias_c}. The shaded green region indicates the case $\alpha\in[\amin,\amax]$, \emph{i.e.}, when no increase of performance is guaranteed by Theorem~\ref{theorem: group dependent prior}.}
    \label{fig:implicit bias and variance}
  \end{figure}

We  can get several  insights from this simplification. First, if both groups are subject to the same amount of bias, $\ba=\bb$, then both $\amin$ and $\amax$ coincide, $\amin=\amax=1/2$. Hence, according to Theorem~\ref{theorem: group dependent prior}, the $\gamma$-rule fairness mechanism in this case is beneficial to the utility of the group-oblivious algorithm for all budgets $\ax \not =1/2$. For $\ax=1/2$, both the $\gamma$-fair group-oblivious algorithm and the group-oblivious algorithm will perform the same $\Qfair{}=\Qgreedy{}$ for all $\gamma >0$.  Hence, if the amount of bias is the same, the result is not different from the one when there is no bias at all (see Section~\ref{ssec:group-independent quality}), which is natural and expected. We illustrate this result in Fig.~\ref{fig:bias_a} which is the same as Fig.~\ref{fig: all 1 stage_a}. 

Second, if the estimate for the high-variance group $A$ has smaller bias than for the low-variance group $B$, i.e., $\db=\ba-\bb <0$, then the $\gamma$-fair mechanism will improve the utility of  the group-oblivious algorithm for all $\ax < 1/2$. It can be seen from the fact that in this case $\Phi\left(\frac{-\db}{\dsx}\right) \ge 1/2$ which means that $\amin=1/2$. This case is illustrated in Fig.~\ref{fig:bias_b}.



Perhaps counterintuitively, when implicit bias and implicit variance both affect the estimation, for some values of $\ax  \in (\amin, \amax)$ specified in Theorem~\ref{theorem: group dependent prior}, the $\gamma$-fair group-oblivious algorithm will  always perform worse than the group-oblivious algorithm.  We observe the corresponding phenomenon on both Fig.~\ref{fig:bias_b} and Fig.~\ref{fig:bias_c} around the values of budgets $\ax=0.6$ and $\ax=0.4$, respectively.

Finally, note that the Bayesian-optimal algorithm as well as the demographic parity (implicitly) remove the biases, hence, the results and discussion from Corollary~\ref{theorem:cost of fairness} can also be applied in this section.

\subsection{Group-dependent latent quality distribution and unbiased estimates}
\label{ssec:group-dependent}

We now assume that there is \emph{no bias} but that the underlying \emph{quality distribution is group-dependent}.
We can also distinguish different cases, when we isolate the effect of group-dependency of the distribution of quality by removing the implicit bias from our consideration. In this case, $\Delta\beta=0$ and the budgets specified in Theorem~\ref{theorem: group dependent prior} can be reformulated as follows:
$$\amin=\min\left\{\Phi\left(\frac{\dmq}{\dsx}\right), \Phi\left(\frac{\dmq}{\dst}\right)\right\},\; \amax=\max\left\{\Phi\left(\frac{\dmq}{\dsx}\right), \Phi\left(\frac{\dmq}{\dst}\right)\right\}.$$

\begin{figure}
  \centering
  \begin{subfigure}{.3\textwidth}
    \includegraphics[width=\linewidth]{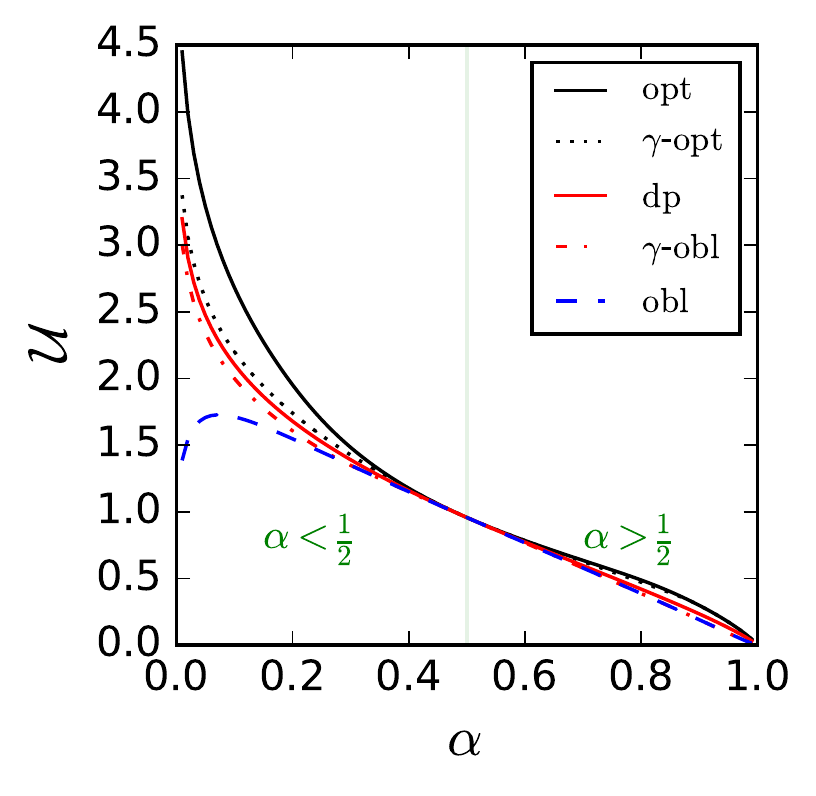}
    \vspace{-.5cm}\caption{$\mqa=\mqb$}
    \label{fig:group-dependent-prior_a}
  \end{subfigure}
  \begin{subfigure}{.3\textwidth}
    \includegraphics[width=\linewidth]{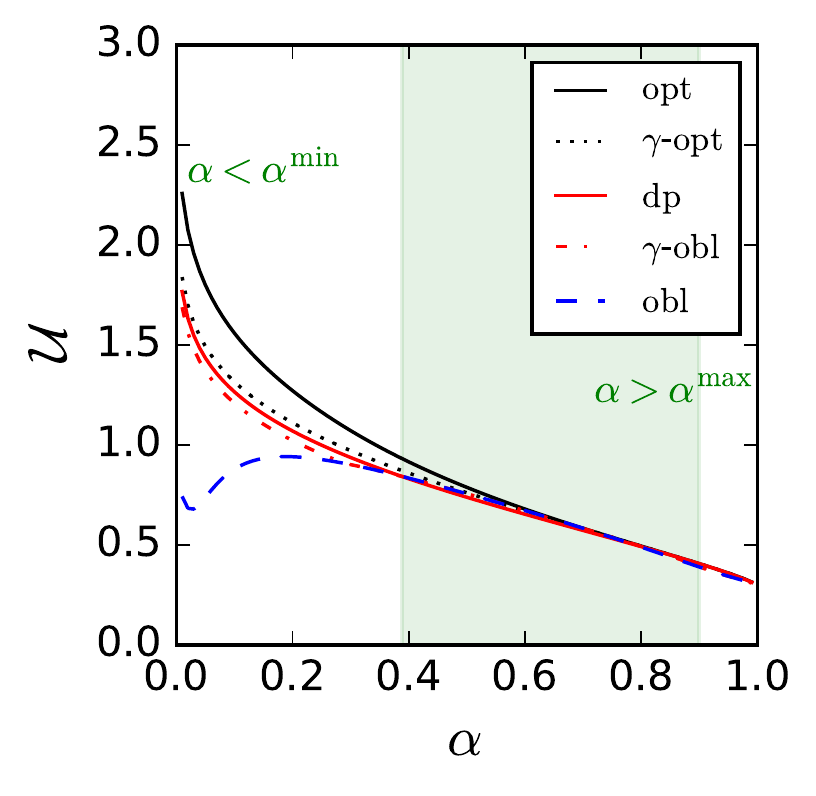} 
    \vspace{-.5cm}\caption{$\mqa < \mqb$}
    \label{fig:group-dependent-prior_b} 
  \end{subfigure}
  \begin{subfigure}{.3\textwidth}
    \includegraphics[width=\linewidth]{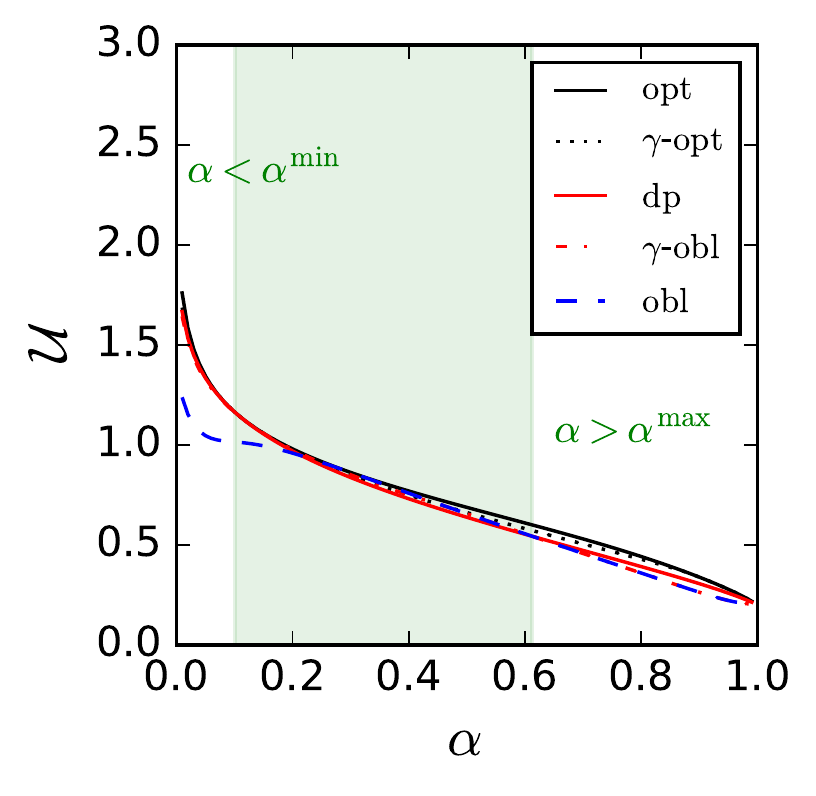} 
    \vspace{-.5cm}\caption{$\mqa > \mqb$}
    \label{fig:group-dependent-prior_c}
  \end{subfigure}
  \caption{The quality of selection in the presence of differential variance for different budgets $\ax$. We assume that $A$-candidates have larger variability of their estimates compare to $B$-candidates $\ssa > \ssb$ as well as the relative amount of noise is larger for $A$-candidates than for $B$-candidates $\sta < \stb$. The implicit variance parameters are equal to $\sxa=3$ and $\sxb=1$. The distribution of quality is $\N(\mqa=0,\sqa=1)$ and $\N(\mqb=0,\sqb=2)$ for \ref{fig:group-dependent-prior_a}, $\N(\mqa=0,\sqa=1)$ and $\N(\mqb=0.5,\sqb=1)$ for \ref{fig:group-dependent-prior_b} and $\N(\mqa=0.5,\sqa=1)$ and $\N(\mqa=0,\sqb=1)$ for \ref{fig:group-dependent-prior_c}. The shaded green region indicates the case $\alpha\in[\amin,\amax]$, \emph{i.e.}, when no increase of performance is guaranteed by Theorem~\ref{theorem: group dependent prior}.}
  \label{fig:different priors}
\end{figure}

We can draw several conclusions from  this simplification. First, if both groups have equal means $\mqa=\mqb$ and if $\ssa > \ssb$, $\sta < \stb$, then the condition in Theorem~\ref{theorem: group dependent prior} simplifies to $\amin=\amax=1/2$, which is equivalent to the result in Corollary~\ref{theorem: 1 stage}. Thus, in this case, the $\gamma$-rule mechanism  improves the quality of group-oblivious selection for all budgets $\ax\not=1/2$.  (If $\ax=1/2$, then $\Qgreedy{}=\Qfair{}$ for all $\gamma$.) We illustrate this case in Fig.~\ref{fig:group-dependent-prior_a} and it is the same result as in Section~\ref{ssec:group-independent quality}.
Second, if both groups have equal variances of quality $\sqa^2=\sqb^2=\sq^2$, then the condition $\sta < \stb$ from Theorem~\ref{theorem: group dependent prior} holds automatically. We  illustrate different cases of relations between $\mqa$ and $\mqb$ in Fig.~\ref{fig:group-dependent-prior_b} and Fig.~\ref{fig:group-dependent-prior_c}.  Unfortunately, the bound on $\Qopt/\Qfopt$ in Theorem~\ref{theorem:cost of fairness general} cannot be further simplified for the case of group-dependent quality distribution. 



\section{Experiments}
\label{section: experiments}

In this section,\footnote{All codes are available at:
  \url{https://gitlab.inria.fr/vemelian/differential-variance-code}
 .} we challenge our theoretical results by using sets of data
that do not satisfy our assumptions. We show in
Section~\ref{ssec:synthetic} that the results are qualitatively
similar when the candidates' true quality comes from a non-normal distribution. 
We
also observe a similar behavior when considering in
Section~\ref{ssec:jee} an artificial scenario that we construct using a real dataset coming from the national Indian exam data. 
We conclude in
Section~\ref{ssec:approximation} with experiments that show that a case with $n=50$ candidates
behaves similarly as with $n=\infty$.

\subsection{Synthetic data with non-normal quality}
\label{ssec:synthetic}

\begin{figure}[ht]
  \centering
  \begin{subfigure}{0.22\textwidth}
    \centering
      \includegraphics[width=\linewidth]{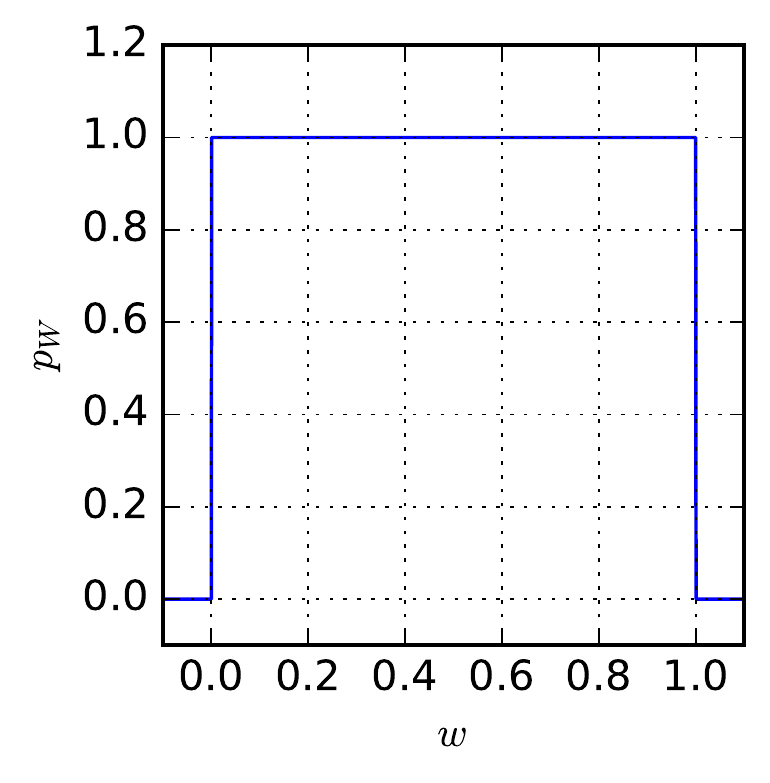}
    \end{subfigure}
    \begin{subfigure}{0.22\textwidth}
      \centering
      \includegraphics[width=\linewidth]{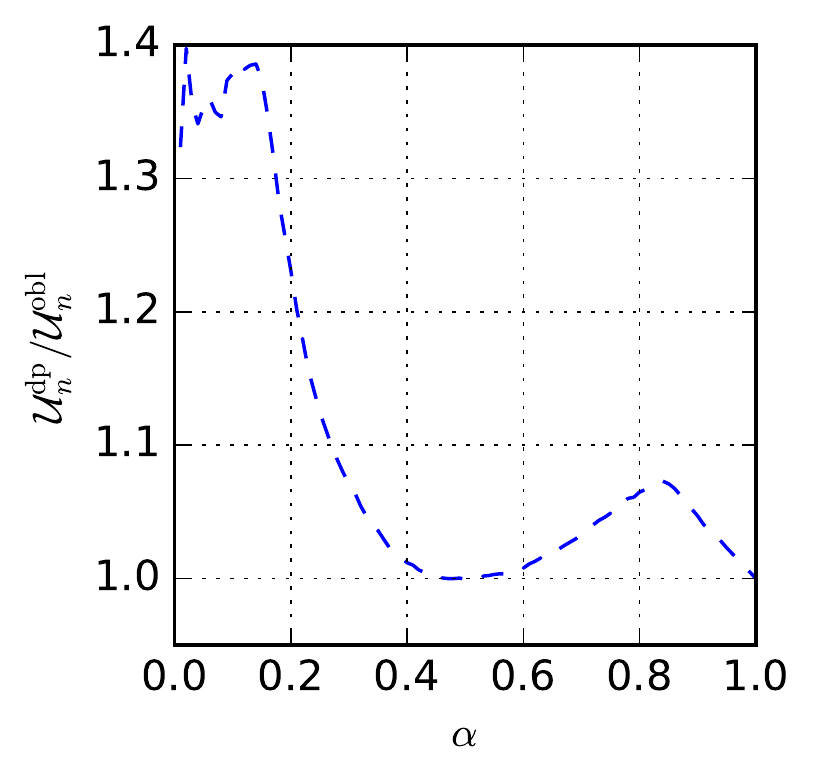}
    \end{subfigure}
    \begin{subfigure}{0.22\textwidth}
      \centering
      \includegraphics[width=\linewidth]{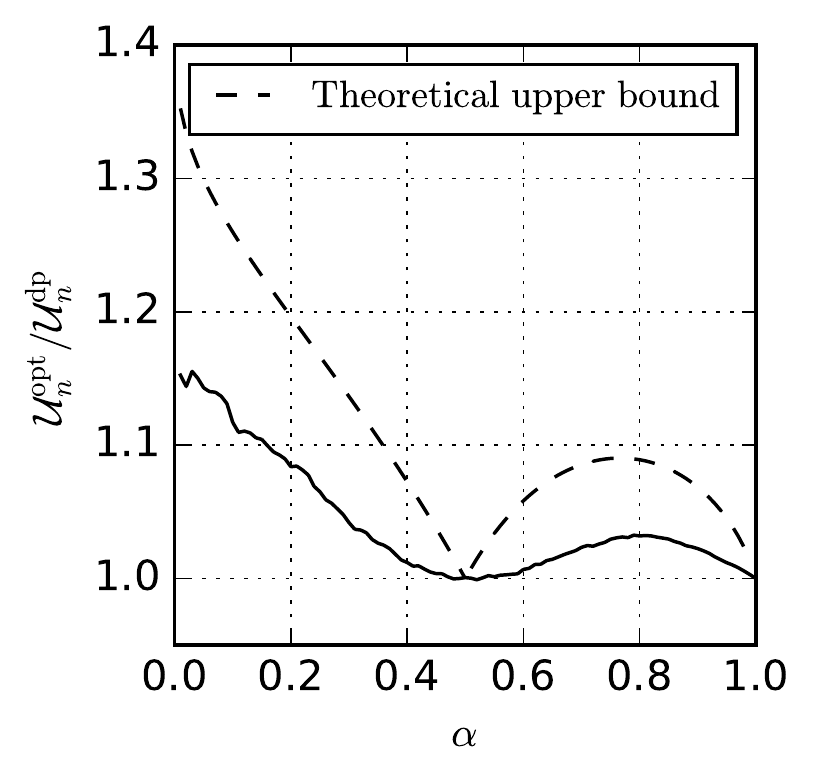}
    \end{subfigure}
    \begin{subfigure}{0.22\textwidth}
      \centering
      \includegraphics[width=\linewidth]{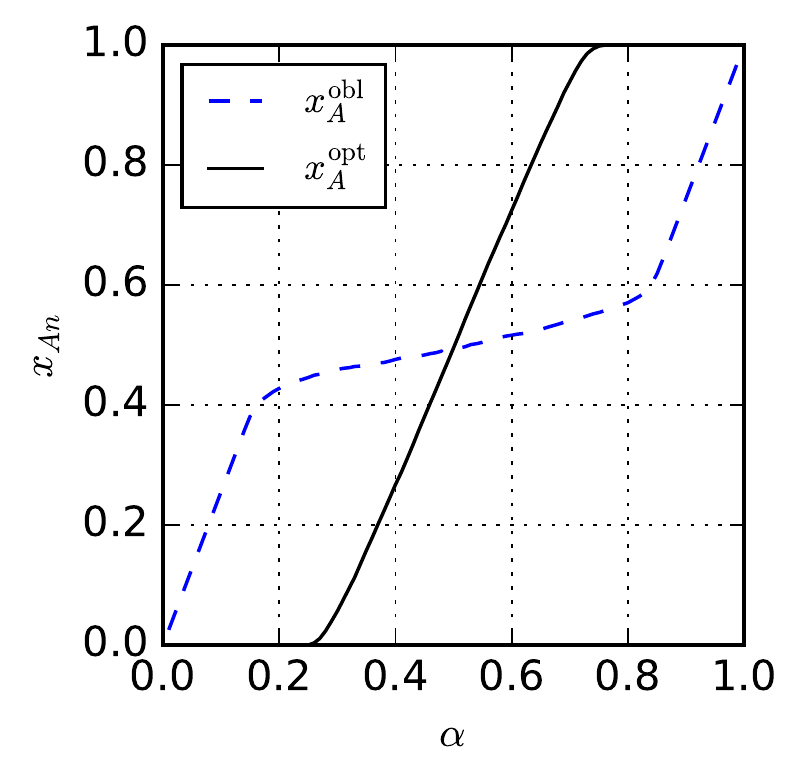}
    \end{subfigure}
    \begin{subfigure}{0.22\textwidth}
      \centering
      \includegraphics[width=\linewidth]{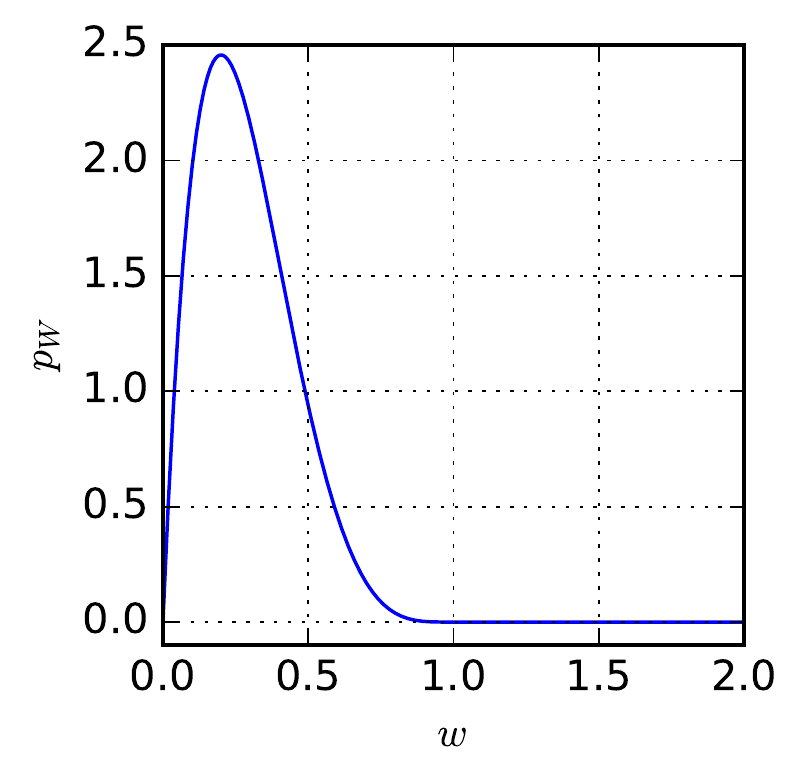}
    \end{subfigure}
  \begin{subfigure}{0.22\textwidth}
    \centering
      \includegraphics[width=\linewidth]{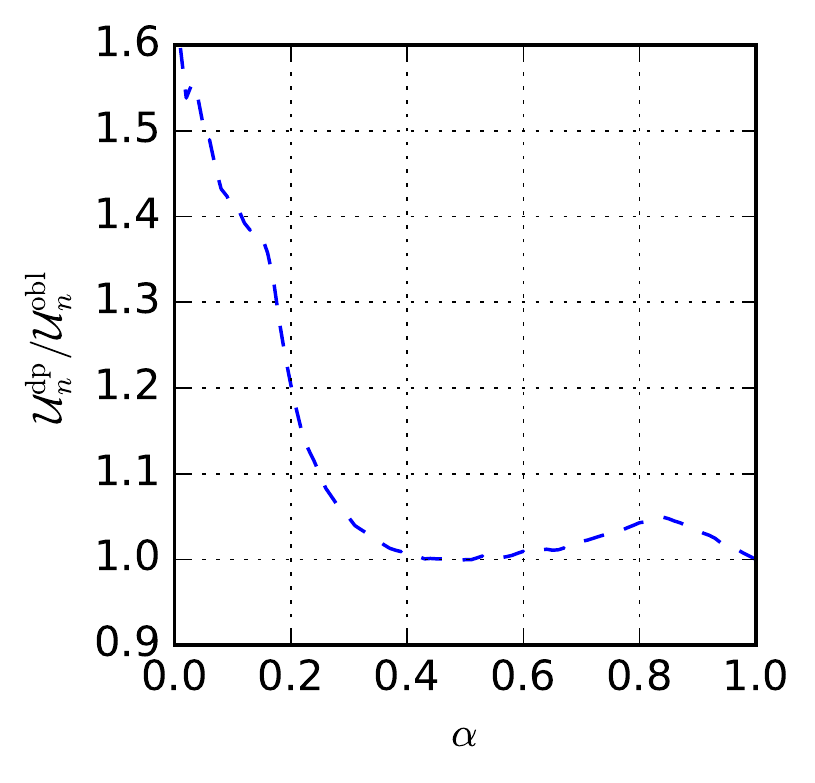}
  \end{subfigure}
  \begin{subfigure}{0.22\textwidth}
    \centering
    \includegraphics[width=\linewidth]{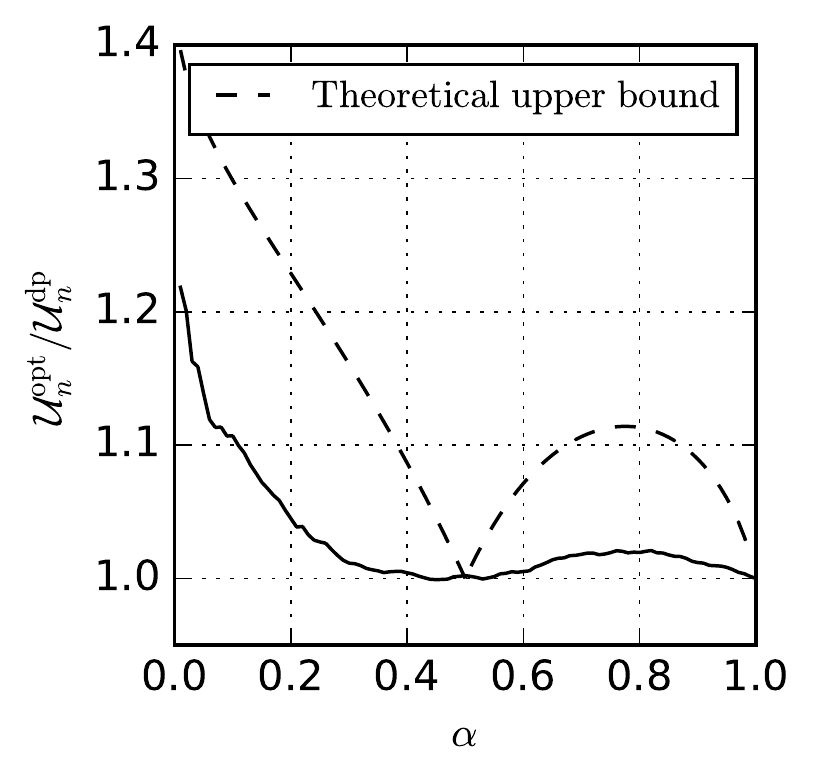}
  \end{subfigure}
  \begin{subfigure}{0.22\textwidth}
    \centering
    \includegraphics[width=\linewidth]{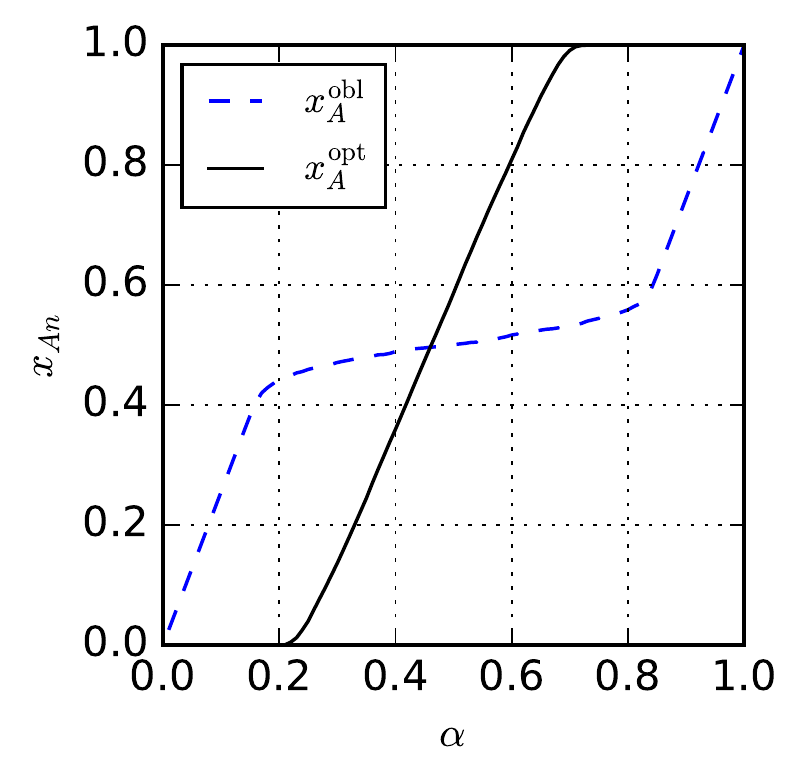}
  \end{subfigure}
  \begin{subfigure}{0.22\textwidth}
    \centering
    \includegraphics[width=\linewidth]{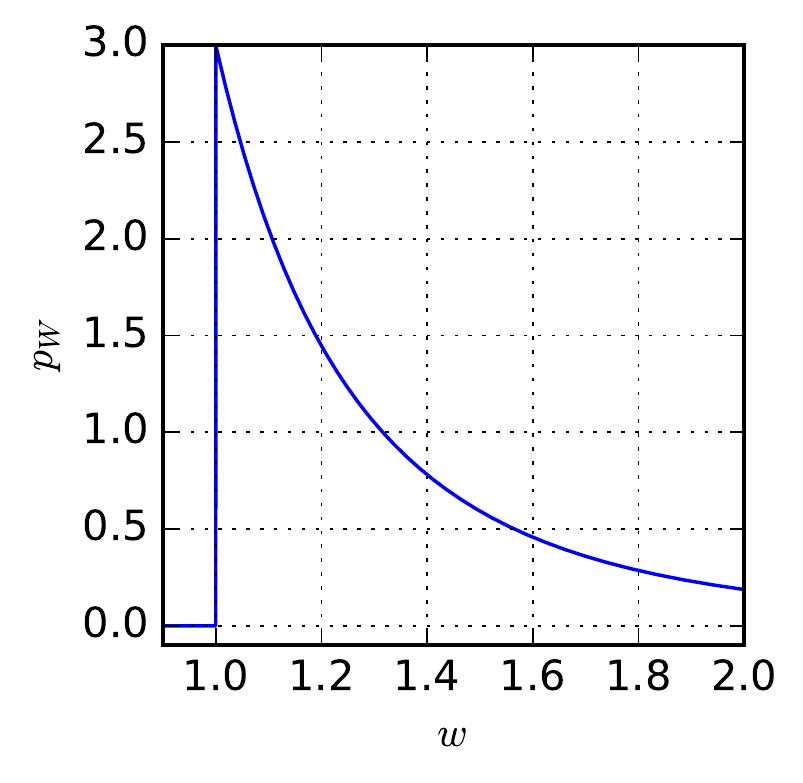}
    \caption{Quality PDF}
    \label{fig:pareto_PDF}
  \end{subfigure}
  \begin{subfigure}{.22\textwidth}
    \includegraphics[width=\linewidth]{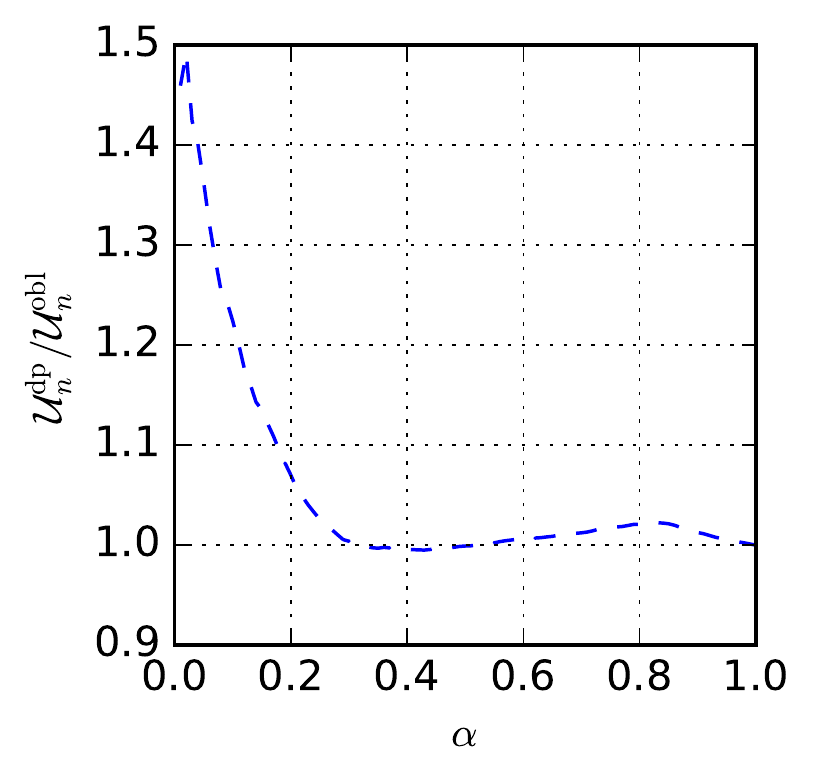}
    \vspace{-.3cm}\caption{Ratio $\Qdp/\Qgreedy{}$}
    \label{fig:pareto_a}
  \end{subfigure}
  \begin{subfigure}{.22\textwidth}
    \includegraphics[width=\linewidth]{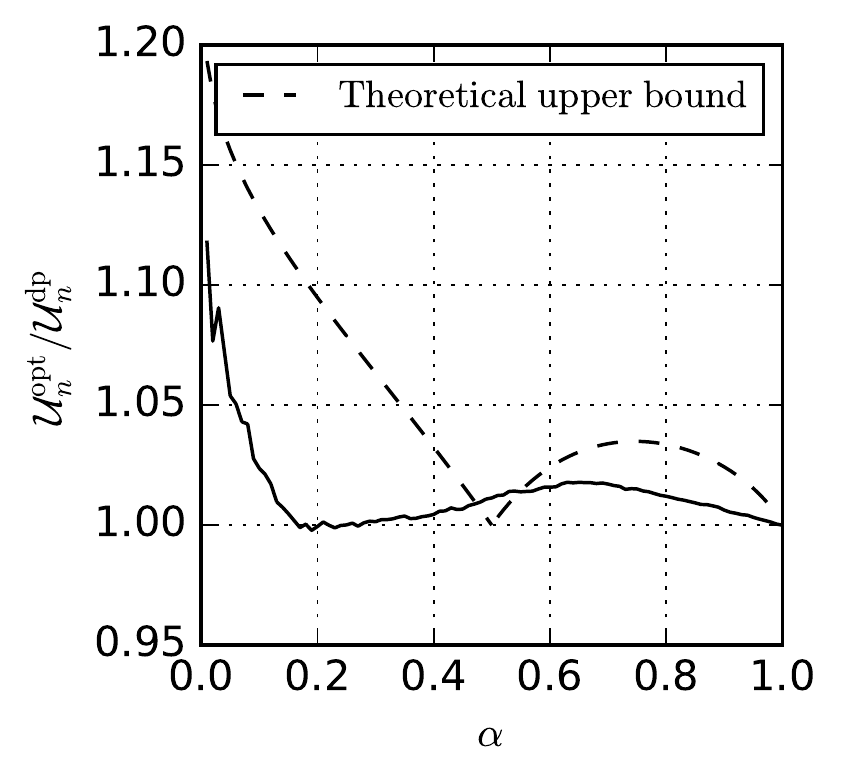}
    \vspace{-.3cm}\caption{Ratio $\Qopt/\Qdp$}
    \label{fig:pareto_b}
  \end{subfigure}
  \begin{subfigure}{.22\textwidth}
    \includegraphics[width=\linewidth]{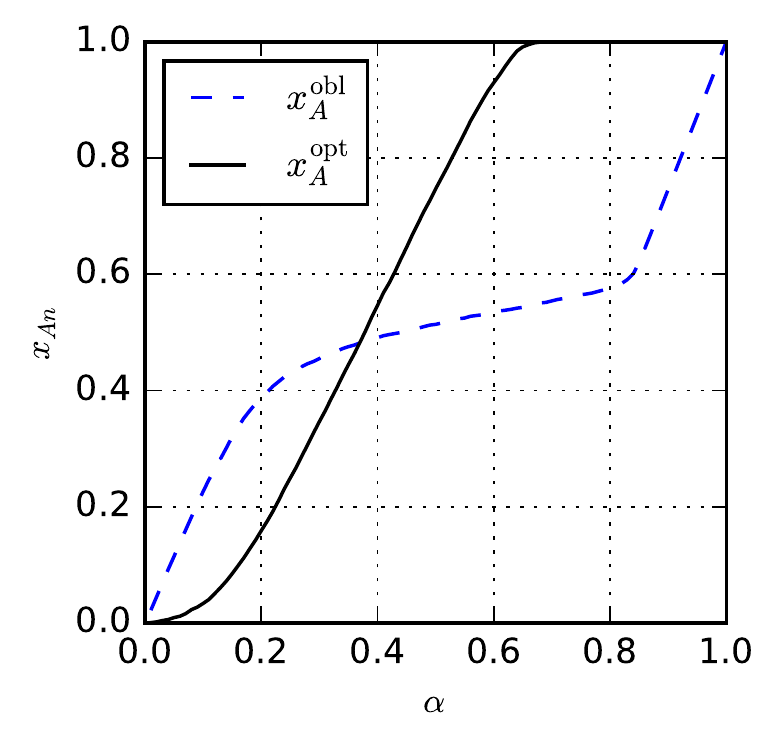} 
    \vspace{-.3cm}\caption{Selection rate $\pxaopt{}$}
    \label{fig:pareto_d}
  \end{subfigure}
  \vspace{-.3cm}\caption{ \textbf{Synthetic data} for different prior distributions of quality $\w$: Uniform on $[0,1]$, Beta(2,5), and Pareto(1,3): Effects of fairness on utility $\Q$ and selection rate $\pxa$. The parameters are $\pa=0.4$, $\sxa=3$ and $\sxb=0.2$. The number of candidates is fixed to $n=10,000$.}
  \label{fig:pareto}
\end{figure}

Our assumption in the theoretical evaluation of Sections~\ref{section: one-stage general case}-\ref{section: cases} was that qualities $\w$
follow a normal distribution. In some cases, however, the quality
distribution is quite different from normal and can be better modeled
by a power law \cite{kleinberg18}, this for example the case for wealth, income or number of citations \cite{Clauset09a}, meaning that a minority possesses a large fraction of the aggregate quality. In this experiment, we vary the quality distribution and consider other distributions of quality $\w$: a Uniform distribution on $[0,1]$, a Beta distribution with the shape parameter equal to $2$ and the scale parameter equal to $5$ or a Pareto distribution with a scale $1$ and shape $3$ (whose PDF $p_\w(w) = \frac{3}{w^{4}}$).  We generate a single dataset of size $n=10,000$. For this dataset we perform a group-oblivious, a demographic parity and a Bayesian-optimal selections. In Fig.~\ref{fig:pareto}, we report the sample utilities $\Q_n$ and sample selection rates $x_{An}$. Note that in this section we consider no bias and group-independent quality distribution. Each line correspond to a different prior quality.

In Fig.~\ref{fig:pareto_a}, we show the performance gap between the group-oblivious and the demographic parity algorithms. We see that the demographic parity improves the utility of the group-oblivious algorithm in most of the cases and that the largest gap
corresponds to the smallest budget $\ax$. Note that contrary to
Corollary~\ref{theorem: 1 stage}, the demographic parity does not always
improve the utility of the group-oblivious algorithm. Yet, the
loss due to the demographic parity is never larger than $0.1\%$ while the
gain can be up to $60\%$.  

In Fig.~\ref{fig:pareto_b}, the performance ratio for the Bayesian-optimal algorithm and the demographic parity algorithm is shown. As expected, the demographic parity harms the utility of the Bayesian-optimal algorithm for both small and large values of budget $\ax$. As the budget $\ax$ increases, the performance gap decreases. To estimate the ratio between the Bayesian-optimal algorithm and the demographic parity, we plot also the value of upper bound from Theorem~\ref{theorem:cost of fairness} that is calculated under an assumption that the quality distribution is normal. We observe that the bound is not tight, however, still dominates the values of $\Qopt/\Qdp$ for most values of budget $\ax$.

Finally, in Fig.~\ref{fig:pareto_d}, we show how the selection fractions $\pxagreedy$ and $\pxaopt$ depend on $\ax$. We see that for small budgets $\ax$, the group-oblivious algorithm tends to select more from group $A$, while for large budgets, the situation is opposite. In contrast, the Bayesian-optimal algorithm always selects $A$-candidates at lower rate if the selection budget $\ax$ is small.



\subsection{IIT-JEE scores dataset}
\label{ssec:jee}

In this section, we aim to consider a scenario in which the underlying quality distributions are non-normal and non-symmetric, and are group-dependent. To easily construct such a case, we create an artificial scenario by using a real dataset, the IIT-JEE dataset \cite{jee09}, with joint entrance exam results in India in 2009. These scores are used as an admission criterion to enter the high-rated universities. The dataset consists of $n=384,977$ records. Every record has information about one student: its name, gender, grade for Mathematics, Physics, Chemistry and total grade. In the dataset, there are 98,028 women and 286,942 men.  This dataset is the same as the one considered in \cite{celis20}.

\begin{figure}
  \centering
  \begin{subfigure}{.24\textwidth}
    \includegraphics[width=\linewidth]{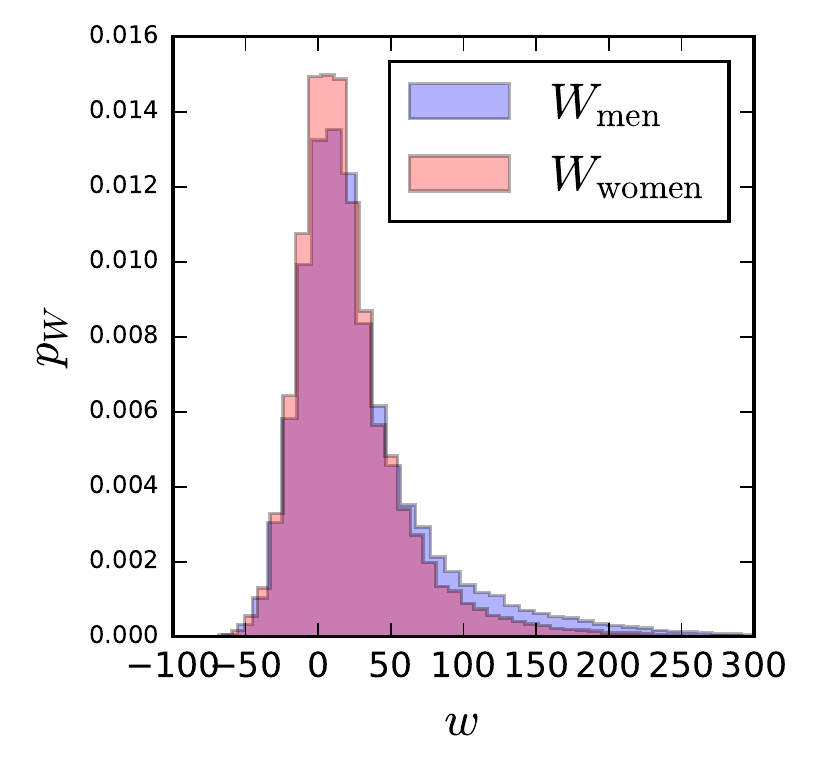}
    \vspace{-.7cm}\caption{Histogram of $\w$}
    \label{fig:jee curves_a}
  \end{subfigure}
  \begin{subfigure}{.24\textwidth}
    \includegraphics[width=\linewidth]{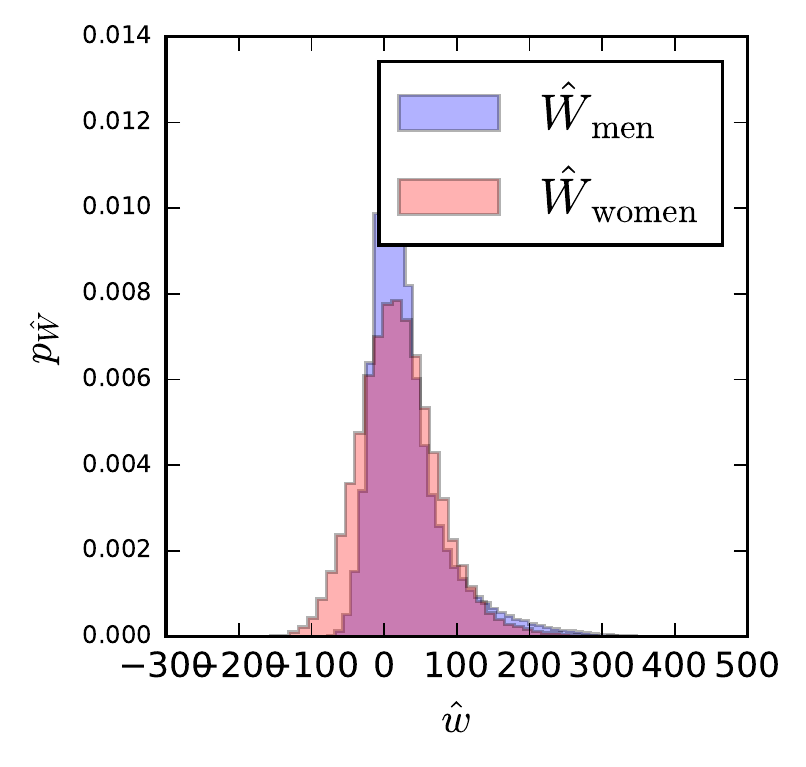}
    \vspace{-.7cm}\caption{Histogram of $\wh$}
    \label{fig:jee curves_b}
  \end{subfigure}
  \begin{subfigure}{.24\textwidth}
    \includegraphics[width=\linewidth]{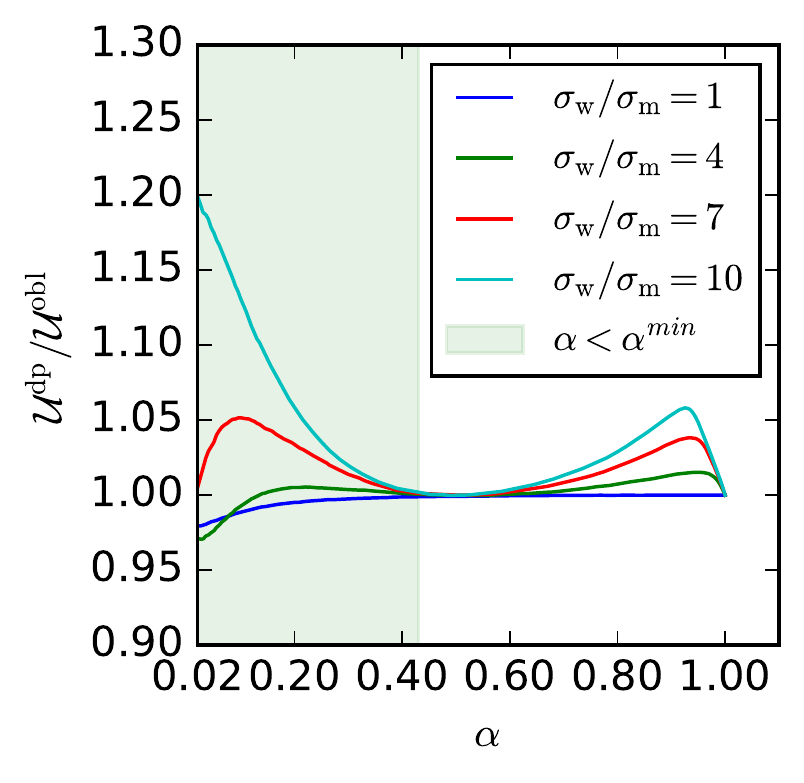} 
    \vspace{-.7cm}\caption{Ratio $\Qdp/\Qgreedy{}$}
    \label{fig:jee curves_c}
  \end{subfigure}
  \begin{subfigure}{.24\textwidth}
    \includegraphics[width=\linewidth]{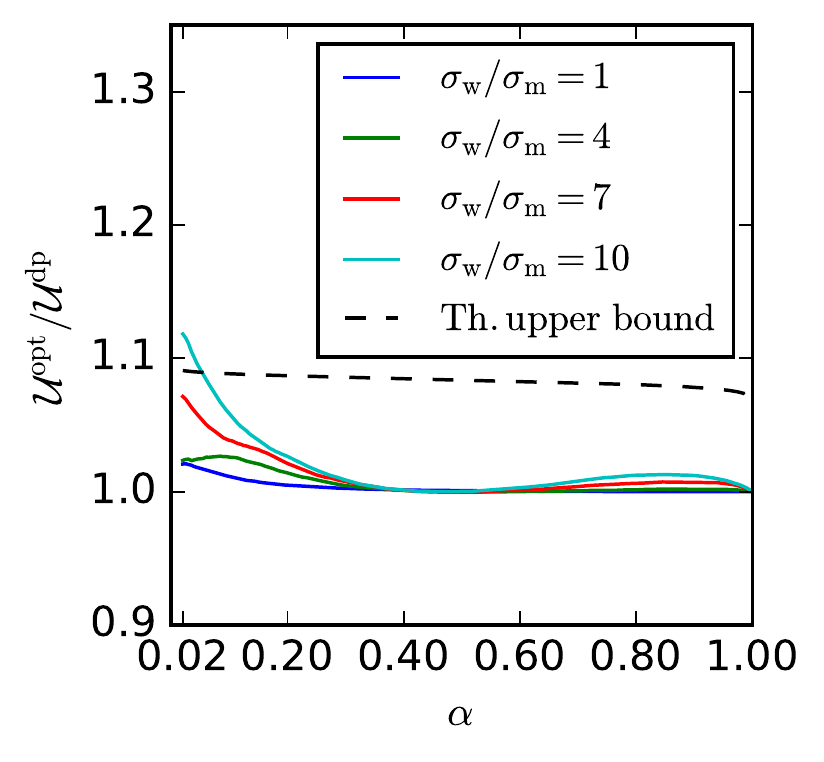}
    \vspace{-.7cm}\caption{Ratio $\Qopt/\Qdp$}
    \label{fig:jee curves_d}
  \end{subfigure}
  \vspace{-.2cm}\caption{   
    Distribution of $\w$ and $\wh$ given gender, and selection ratios for \textbf{IIT-JEE dataset} \cite{jee09}. Mean values and standards deviations of $W$ for two groups are: $\mu_{\mathrm{m}} = 30.8$, $\eta_{\mathrm{m}} = 51.8$,  $\mu_{\mathrm{w}} = 21.2$, $\eta_{\mathrm{w}} = 39.3$. Added noise has standard deviation $\sigma_\mathrm{m}=10$ and $\sigma_\mathrm{w} = k\cdot\sigma_\mathrm{m}$; $k=4$ in plot (b).}
  \label{fig:jee curves}
  \vspace{-.2cm}
\end{figure}

In order to construct a model of differential variance, we consider an artificial scenario where the field ``grade'' is the true latent quality $\w$ of the candidates. The mean values and standard deviations of $W$ for the two groups are: $\mu_{\mathrm{men}} = 30.8$, $\eta_{\mathrm{men}} = 51.8$, $\mu_{\mathrm{women}} = 21.2$, $\eta_{\mathrm{women}} = 39.3$. We then suppose that an unbiased estimator $\wh$ of the grade is observed. The standard deviation of estimation for male candidates is set to $\sigma_\mathrm{men} = 10$. For the women group, which is the minority group, we consider different cases: $\sigma_\mathrm{women} = k\cdot\sigma_\mathrm{men}$, for $k=1,4,7,10$.  The distribution of grades $\w$ and observed values $\wh$ for $k=4$ are shown in Fig.~\ref{fig:jee curves_a} and \ref{fig:jee curves_b}.


For the dataset we perform a group-oblivious (select best $m$), a demographic parity selection (select best $m$, but maintain the demographic parity condition $\pxa=\pxb$ up to one candidate) and a Bayesian-optimal selection. The selection size varies from  2\% to 100\% of total number of candidates, i.e., out of 384,977 students the decision maker selects 7,700 students or more. A selection rate of 2\% was set by IIT in 2009 \cite{celis20}.

The results for the ratio of $\Qdp/\Qgreedy$ are given in Fig.~\ref{fig:jee curves_c}. We observe that for both small and large values of $\ax$, the demographic parity helps the utility of the group-oblivious algorithm, if the noise values of women evaluation $\sigma_\mathrm{women}$ are large, which agrees with the results from Theorem~\ref{theorem: group dependent prior}. We see that the gain can be up to around 20\% if the selection size is small and up to 5\% if the selection size is large. For the case where $\sigma_\mathrm{women}$ and $\sigma_\mathrm{men}$ are close, we observe no gain if the selection is large and we observe a minor loss in utility (around 2\%) if the selection is small. This is due to the fact that in the dataset, there are more men with a high true latent quality $\w$, as seen in Fig.~\ref{fig:jee curves_a}. We also plot the region (for $k=10$) from Theorem~\ref{theorem: group dependent prior} in which the utility of the demographic parity algorithm should dominate the utility of the group-oblivious algorithm if the distribution of quality is  a group-dependent normal.

Finally, on Fig.~\ref{fig:jee curves_d} we show the ratio $\Qopt/\Qdp$ for different values of $k=1,4,7,10$. In addition to these ratios, we also plot the bound  from Theorem~\ref{theorem:cost of fairness general} for $k=10$. We see that the bound is quite close to the actual value of $\Qopt/\Qdp$ for small $\ax$.

\subsection{Accuracy of the approximation for small $n$}
\label{ssec:approximation}

As discussed in Section~\ref{section: model}, we cannot solve the problem with finite selection sizes exactly. Instead, throughout the paper, we use an approximation that is exact as number of candidates $n$ tends to infinity (Theorem~\ref{th:n_infinity}). In this section, we question the accuracy of this approximation when the number $n$ of candidates is relatively small. For our experiment, we generate datasets of different sizes $n=50, 100$. For every size parameter $n$, we generate $10,000$ different datasets. For a population size $n$, we denote by $\langle\Q_n\rangle$ the average quality of the selected candidates over our $10,000$ experiments. In each case, the true latent qualities $\w$ are generated from a normal distribution $\N(1,1)$.

\begin{figure}
  \centering
  \begin{subfigure}{.32\textwidth}
    \includegraphics[height=.95\linewidth]{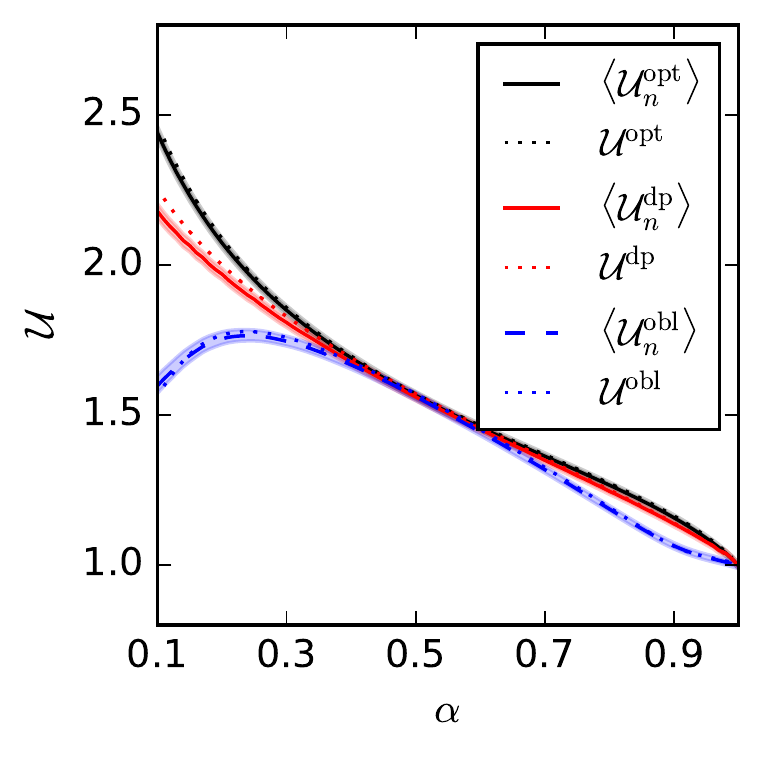} 
    \vspace{-.3cm}\caption{Average utility $\langle \Q_n\rangle$, $n=100$}
    \label{fig: finite normal_a}
  \end{subfigure}
  \begin{subfigure}{.32\textwidth}
    \includegraphics[height=.95\linewidth]{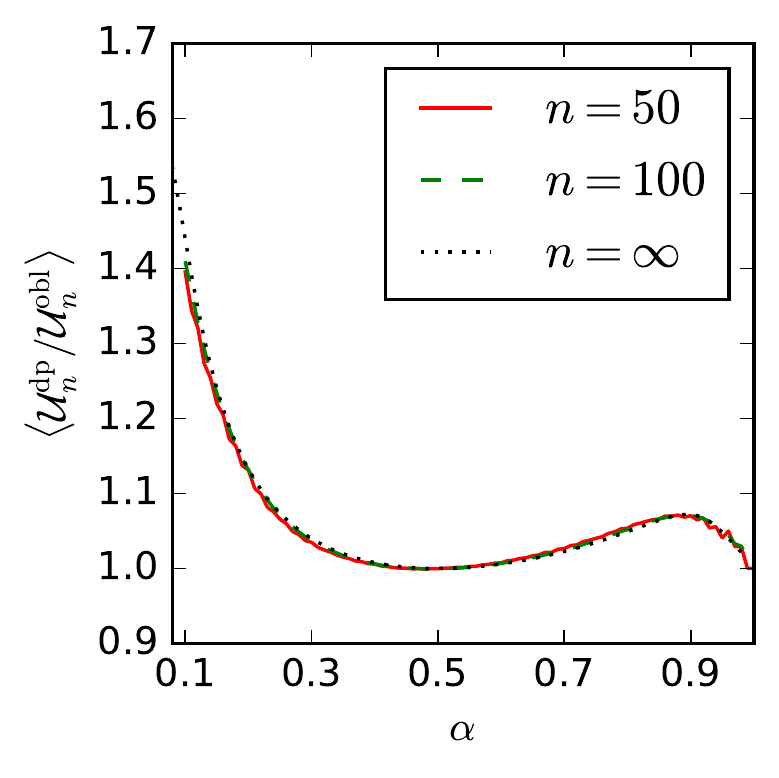}
    \vspace{-.3cm}\caption{Ratio $\langle \Qdp_n/\Qgreedy_n\rangle$}
    \label{fig: finite normal_b}
  \end{subfigure}
  \begin{subfigure}{.32\textwidth}
    \includegraphics[height=.95\linewidth]{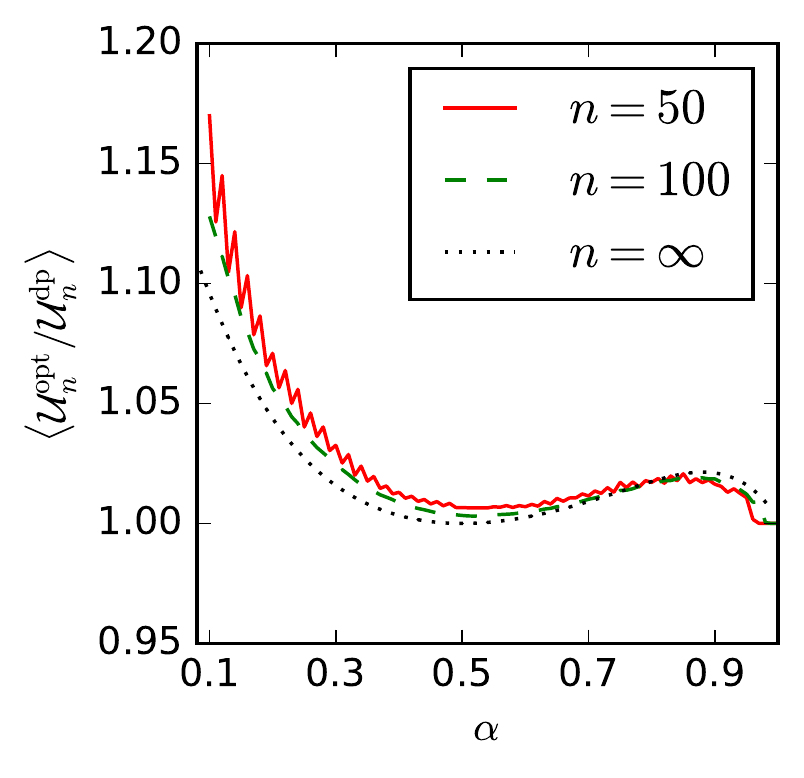}
    \vspace{-.3cm}\caption{Ratio $\langle \Qopt_n/\Qdp_n\rangle$}
    \label{fig: finite normal_c}
  \end{subfigure}
  \vspace{-.2cm}\caption{\textbf{Finite population size}: quality of selection and expected gain of the demographic parity over the group-oblivious algorithm. The quality distribution $\w$ is $\N(\mq=1,\sq^2=1)$ and the noise parameters are $\sxa = 3$, $\sxb=0.2$. The number of experiments per set of parameters is $K=10,000$. The shaded areas are the confidence intervals (corresponding to one standard deviation on the estimation of the empirical mean).}
  \label{fig: finite normal}
\end{figure}

In Fig.~\ref{fig: finite normal_a} we plot the average utilities
$\langle\Q_n\rangle$ for a population of $n=100$, where we select
$m$ individuals and where we vary $m$ from 10 to 100. The
shaded region corresponds to a confidence interval.  We consider three
selection algorithms (demographic parity, group-oblivious and Bayesian-optimal) and
compare the performance for $n=100$ with the limiting quantities
$\Qdp$, $\Qgreedy$ and $\Qopt$.  We observe that, even for $n=100$, the average
values of utility are close to the approximation.  In Fig.~\ref{fig:
  finite normal_b} we compare the average ratio of performances $\langle
\Qdp_n/\Qgreedy_n\rangle$ for different $n$. We observe
that the approximation for $n=50$ is a good prediction of the average
gain provided by the use of demographic parity. Similarly, in Fig.~\ref{fig: finite normal_c}, we compare the average ratio of performances $\langle
\Qopt_n/\Qdp_n\rangle$ for different $n$. Again, the curves for finite $n$ are still quite close to the case where $n\to\infty$.


\section{Conclusion}
\label{section: discussion}

In this work, we study a simple model of the selection problem that captures the phenomenon of differential variance, that is, the decision maker has estimates of the candidates' quality with different variances for different demographic groups. We distinguish two notable cases. In the first case, the decision maker does not have information about the estimate properties (variances and biases); as a result they use a group-oblivious algorithm. In the second case, every information about the distribution of quality is known, and the decision maker is Bayesian-optimal.

First, we show that both baseline algorithms (without any fairness constraint) lead to discrimination. Then we identify conditions under which, in the first case, the $\gamma$-rule fairness mechanism (a generalization of the $\ff$ rule) leads to a higher selection utility compared to using the group-oblivious baseline. In the second setting, the $\gamma$-rule mechanism is harmful to the utility of Bayesian-optimal baseline but we prove that the utility decrease is bounded. Overall, our results contribute to a recent thread of works identifying cases in which, contrary to conventional wisdom, imposing fairness mechanisms does not come at the cost of utility (or even if it does, that the cost is bounded). 
Beyond fitting a particular application in detail, our results are useful in thinking about the impact of possible policies. For instance, they can help evaluate the effect of imposing a given fairness mechanism, or deciding whether or not to allow access to group information in a particular application.



Our theoretical results are obtained under the assumption that the true latent quality $\w$ follows a normal law (to allow for analytical derivations). This assumption can be relaxed: we can plug into the model any distribution of latent quality (e.g., Pareto, uniform, etc.). We show numerically in Section~\ref{section: experiments} that it does not change the flavor of the main results. Extending these results theoretically is, however, challenging as in our proofs we operate with the expression for the conditional expectation of true latent quality given the noisy estimate. In a non-normal case, this conditional expectation cannot, in general, be expressed in closed form, which complicates the analysis.

Our modeling assumptions imply that a candidate's quality does not depend on the selection strategy used. If attaining a certain level of quality comes at a cost, then the interaction between decision makers and candidates may be seen as a game. It would be interesting to see how differential variance affects the incentives of candidates and how the $\gamma$-rule changes them in this game. We leave this game-theoretic formulation of the selection problem as a future direction.

\section*{Acknowledgments}

This work has been partially supported by MIAI @ Grenoble Alpes (ANR-19-P3IA-0003), by the French National Research Agency (ANR) through grant ANR-20-CE23-0007, and by a European Research Council (ERC) Advanced Grant for the project ``Foundations for Fair Social Computing'' funded under the European Union's Horizon 2020 Framework Programme (grant agreement no. 789373). We thank the editor and the reviewers for their thoughtful comments.

\bibliographystyle{elsarticle-num}
\bibliography{bibliography} 

\appendix

\section{Ommited proofs}
\label{section: proofs}

In this section we provide detailed proofs of the statements given before. Namely, these are proofs of Theorem~\ref{theorem: group oblivious general case}, Theorem~\ref{theorem: bayesian optimal general case}, Theorem~\ref{theorem: group dependent prior} and Theorem~\ref{theorem:cost of fairness general}.

\subsection{Proof of Theorem~\ref{theorem: group oblivious general case}}
\label{proof:group oblivious general case}
  By our assumptions, the estimates of qualities for $G$-candidates follow  a normal law with the mean $\mqg-\bg$ and the variance $\ssg^2 = \sqg^2 + \sxg^2$.  Recall that the selection rate $\pxggreedy{}$ for the group-oblivious algorithm is a probability for the $G$-candidate to have an estimated quality larger than a predefined group-independent threshold: $\pxggreedy{ = }\Pb(\wh \ge \txggreedy|G)$. 
  Taking all that into account, the selection rate for $G$-candidates can be expressed as:
  \begin{align*}
    \pxggreedy{}=1-\Phi\left(\frac{\txggreedy-\mqg + \bg}{\ssg}\right).
  \end{align*}
  This shows that the condition $\pxagreedy{} > \pxbgreedy{}$ is equivalent to  $\frac{\txggreedy - \mqa + \beta_A}{\ssa}  <  \frac{\txggreedy - \mqb + \beta_B}{\ssb}$, since $\Phi$ is an increasing function of its argument. Hence, by rearranging the terms we conclude that $\txggreedy > \frac{\mqa\ssb -\mqb\ssa}{\ssb-\ssa} + \frac{\beta_B\ssa -\beta_A\ssb}{\ssb - 
  \ssa}$. By substituting the corresponding threshold to the expression for the selection rate $\pxggreedy{}$, we end up with the expression for the values of budgets $\ax$ for which $\pxagreedy{} > \pxbgreedy{}$. The calculations show that for the budgets $\ax<1-\Phi\left( \frac{(\mqa  - \mqb) - (\beta_A - \beta_B)}{\ssb - \ssa}\right)=\Phi\left(\frac{\Delta\mq - \Delta\beta}{\dsx}\right)$ using the group-oblivious algorithm leads to overrepresentation of a high-variance group $A$, where we use the notation $\Delta\mq=\mqa-\mqb$, $\Delta\beta=\ba-\bb$ and $\dsx=\ssa - \ssb$.

\subsection{Proof of Theorem~\ref{theorem: bayesian optimal general case}}
\label{proof:bayesian optimal general case}

The Bayesian-optimal algorithm selects candidates for which the expected quality $\wt$ is larger than some group-independent but budget-dependent threshold $\tilde \theta$. Since $\wtg$ follows a normal law with the mean $\mqg$ and the variance $\stg^2$, we can write  that
$\pxgopt{} = 1- \Phi\left(\frac{\tilde \theta - \mqg}{\stg}
\right).$ 
In the rest of the proof, without loss of generality we assume that $\sta^2 < \stb^2$, hence, we can calculate that $$\pxaopt < \pxbopt \iff \frac{\tilde \theta - \mqa}{\sta} > \frac{\tilde \theta - \mqb}{\stb} \iff \tilde \theta > \frac{\mqa\stb -\mqb\sta}{\stb-\sta}.$$
By substituting the corresponding threshold to the expression for the selection rate $\pxgopt$, we end up with the expression for the values of budgets $\ax$ for which $\pxaopt < \pxbopt$. The calculations show that  this is for all budgets $\ax<\Phi\left(\frac{\dmq }{\dst}\right)$, where we use the notation $\dmq=\mqa-\mqb$ and $\dst=\sta - \stb$.

\subsection{Properties of the utility $\Q$ (Proof of Theorem~\ref{theorem: group dependent prior})}
\label{ssec: properties of utility}

In this section, we study the properties of the utility function $\Q(\pxa)$ independently from the selection algorithm used. We give the expression for the derivative of $\Q$ as a function of $\pxa$. This expression allows us to prove that the utility function $\Q$ is strictly concave. This implies that as we have $\pxagreedy \le \pxafair \le \pxafopt \le \pxaopt$ or $\pxagreedy \le \pxafair \le \pxafopt \le \pxaopt$, one always has $\Q(\pxagreedy) \le \Q(\pxafair) \le \Q(\pxafopt) \le \Q(\pxaopt)$, with strict inequalities whenever the above inequalities are strict.

\begin{lemma}
  \label{lemma:U is convave}
Assume that the budget $\ax$ is fixed. 
\begin{enumerate}
  \item The first derivative of the utility $\Q(\pxa)$ can be expressed as follows: 
  \begin{align}
    \label{eq: Q1'}
    \Q'(\pxa) = \frac{\pa}{\ax}\left[ \frac{(\txa + \ba)\sqa^2 + \mqa\sxa^2}{\sqa^2 + \sxa^2} - \frac{(\txb + \bb) \sqb^2 + \mqb\sxb^2}{\sqb^2 + \sxb^2}\right]
  \end{align}
  where $\twha, \twhb$ are such that $\Pb(\wh \ge \twha\,|\,G=A)=\pxa$ and $\sum_{G\in\{A,B\}}\Pb(\wh \ge \twhg\,|\,G)\cdot\pg = \ax$.
\item The utility $\Q(\pxa)$ is strictly concave.
\end{enumerate}
\end{lemma}

\begin{proof}
  By definition of $\Q$ in \eqref{eq: u}, the utility $\Q$ equals $\V(\twha, \twhb)$ where $\twha, \twhb$ are the unique thresholds such that $\Pb(\wh \ge \twha\,|\,G=A) = \pxa$ and $\sum_{G\in\{A,B\}}\Pb(\wh \ge \twhg\,|\,G)\cdot\pg = \ax$.  Using that $\wtg$ and $\whg$ are normally distributed these quantities can be  expressed as:
\begin{align*}
&\V(\twha, \twhb) = \frac 1 {\ax} \sum_G \pg  \int_{\txg}^{\infty} d\hat w \int_{-\infty}^{\infty} dw \, \left[w\cdot \frac{1}{\sqg}\phi\left(\frac{w-\mqg}{\sqg}\right) \cdot \phig \right],\\
&\pxg(\twhg)=  \int_{\txg}^{\infty} d\hat w \int_{-\infty}^{\infty}  dw\, \left[\frac{1}{\sqg}\phi\left(\frac{w-\mqg}{\sqg}\right) \cdot \phig\right].
\end{align*}

Using the chain rule, we can write the first derivative of selection utility:
\begin{align}
\frac{d\Q}{d\pxa} &= \sum_G \frac{\partial \V}{\partial \twhg}\frac{d\twhg}{d\pxa}.
\label{eq: derivative}
\end{align}
From the budget constraint $\pa\pxa + \pb\pxb = \ax$, by differentiating both parts by $\pxa$, we obtain that
$\pa \frac{d\pxa}{d\pxa} + \pb\frac{\partial\pxb}{\partial\twhb} \frac{d\twhb}{d\pxa} = 0$
which implies that $\frac{d\twhb}{d\pxa} = -\frac{\pa}{\pb} \frac{\partial \twhb}{\partial \pxb}.
$
Then, by substituting the obtained expression for $\frac{d\twhb}{d\pxa}$ into \eqref{eq: derivative}, we obtain that $\frac{d\Q}{d\pxa} = \pa \left( \frac{\partial \V}{\partial \twha}\frac{\partial\twha}{\partial\pxa}- \frac{\partial \V}{\partial \twhb}\frac{\partial\twhb}{\partial\pxb} \right)$.  From this, the expression  \eqref{eq: Q1'}  follows directly.
 
We  observe that the first derivative is linear in the selection thresholds $\txa$ and $\txb$. Thus, as the selection rate $\pxa$ increases, the derivative $\Q'_{\pxa}$ decreases which means that the  function $\Q(\pxa)$ is strictly concave.
\end{proof}

\subsection{Proof of Theorem~\ref{theorem:cost of fairness general}}
\label{proof:cost of fairness}
  Assume that $\ax < \Phi\left(\frac{\dmq}{\dst}\right)$. By Theorem~\ref{theorem: bayesian optimal general case}, we have $\pxaopt{} < \pxadp{}$. As we prove in Lemma~\ref{lemma:U is convave}, the utility function $\Q$ is concave function of $\pxa$. Using the concavity of $\Q$ we have $\frac{\Q(\pxaopt) - \Q(\pxadp)}{\pxaopt - \pxadp} \ge \Q'(\pxa=\pxadp)$ which implies that $\Q(\pxaopt) - \Q(\pxadp) \le (0-\ax) \cdot \Q'(\pxadp)$, where we use the fact that $\pxadp - \pxaopt  \le \ax$ for all budgets $\ax < \Phi\left(\frac{\dmq}{\dst}\right)$.

 By dividing both sides by $\Q(\pxadp)$, from the above inequality we obtain the following upper bound:
  \begin{align*}
    \frac{\Q(\pxaopt)}{\Q(\pxadp)} \le 1 - \ax\cdot\frac {\Q'(\pxadp)}{\Q(\pxadp)},
  \end{align*}
The expression for $\Q'(\pxadp)$ can be written explicitly by using \eqref{eq: Q1'} and the fact that group-dependent thresholds  for the demographic parity algorithm can be calculated as $\txgdp = \mqg -\bg + \ssg\Phi^{-1}(1-\ax)$:
  \begin{align*}
    \Q'(\pxadp) = \frac \pa{\ax}\left( \mqa - \mqb + \Phi^{-1}(1-\ax)\left[\frac{\sqa^2}{\sqrt{\sxa^2+\sqa^2}} - \frac{\sqb^2}{\sqrt{\sxb^2+\sqb^2}} \right]\right)
    = \frac{\pa}{\ax}\left( \dmq + \Phi^{-1}(1-\ax)\dst\right).
  \end{align*}
  The utility by the demographic parity algorithm can be calculated using the law of total expectation and the expected value of truncated normal distribution as follows:
  \begin{align*}
    \Q(\pxadp) = \sum_G \pg \mqg + \frac{\phi\left(\Phi^{-1}\left(1-\ax\right) \right)}{\ax}\sum_G \pg \frac{\sqg^2}{\sqrt{\sxg^2+\sqg^2}} = \sum_G \pg \mqg + \frac{\phi\left(\Phi^{-1}\left(1-\ax\right) \right)}{\ax}\sum_G \pg \stg.
  \end{align*}

   Hence, from the above inequality and the expressions for $\Q(\pxadp)$ and $\Q'(\pxadp)$, we can obtain the following upper bound on the ratio $\Qopt/\Qdp$ for $\ax < \Phi(\dmq/\dst)$\footnote{{Note that the case $\ax > \Phi(\dmq/\dst)$ is proven similarly, except that we use $\pxaopt  - \pxadp \le 1 - \ax$.}}:
  \begin{align*}
    \frac{\Q(\pxaopt)}{\Q(\pxadp)} &\le 1 - \ax \cdot\frac\pa\ax\frac{\dmq +  \Phi^{-1}(1-\ax)\dst}{\sum_G \pg \mqg + \frac{\phi\left(\Phi^{-1}\left(1-\ax\right) \right)}{\ax}\sum_G \pg \stg}.
  \end{align*}

  For the  $\gamma$-fair Bayesian-optimal algorithm, the upper bound on $\Qopt/\Qdp$  can be calculated in a similar manner. The values of the selection rate difference for $\ax < \Phi(\dmq/\dst)$ can be upper bounded as $\pxafopt{} - \pxaopt  \le \frac{\ax}{\pa + \pb/\gamma}$, since the selection by the Bayesian-optimal algorithm lies either inside the $\gamma$-region $\pxa \in \left[\frac{\ax}{\pa + \pb/\gamma}, \frac{\ax}{\pa + \gamma\pb}\right]$ or on its boundary. For $\ax > \Phi(\dmq/\dst)$, the difference can be upper bounded as $\pxaopt{} - \pxafopt  \le 1 - \frac{\ax}{\pa + \gamma\pb}$.







\end{document}